\documentclass[sigconf, screen, nonacm]{acmart}
\usepackage[table]{xcolor}
\usepackage{multirow}
\usepackage[ruled,vlined]{algorithm2e}

\AtBeginDocument{\providecommand\BibTeX{{Bib\TeX}}}

\definecolor{ltgray}{gray}{0.93}
\newcommand{\best}[1]{\textcolor{red}{#1}}
\newcommand{\second}[1]{\textcolor{blue}{#1}}
\newcommand{\method}{PACO}
\SetKwInput{KwIn}{Input}
\SetKwInput{KwOut}{Output}
\SetKwInput{KwInit}{Init}

\makeatletter
\@ACM@balancefalse
\makeatother

\begin{document}

\title{PACO: Proxy-Task Alignment and Online Calibration for On-the-Fly Category Discovery}

\author{Weidong Tang}
\authornote{These authors contributed equally to this work.}
\email{wdtang29@gmail.com}
\affiliation{
  \institution{China Agricultural University}
  \department{College of Information and Electrical Engineering}
  \city{Beijing}
  \country{China}}

\author{Bohan Zhang}
\authornotemark[1]
\email{zbohan082@gmail.com}
\affiliation{
  \institution{China Agricultural University}
  \department{College of Information and Electrical Engineering}
  \city{Beijing}
  \country{China}}

\author{Zhixiang Chi}
\email{zhxchi@ece.utoronto.ca}
\affiliation{
  \institution{University of Toronto}
  \department{Department of Electrical and Computer Engineering}
  \city{Toronto}
  \country{Canada}}

\author{ZiZhang Wu}
\email{wuzizhang87@gmail.com}
\affiliation{
  \institution{Fudan University}
  \city{Shanghai}
  \country{China}}

\author{Yang Wang}
\email{yang.wang@concordia.ca}
\affiliation{
  \institution{Concordia University}
  \department{Department of Computer Science and Software Engineering}
  \city{Montreal}
  \country{Canada}}

\author{Yanan Wu}
\authornote{Corresponding author.}
\email{ynwu@cau.edu.cn}
\affiliation{
  \institution{China Agricultural University}
  \department{College of Information and Electrical Engineering}
  \city{Beijing}
  \country{China}}

\renewcommand{\shortauthors}{Tang et al.}

\begin{abstract}
On-the-Fly Category Discovery (OCD) requires a model, trained on an offline support set, to recognize known classes while discovering new ones from an online streaming sequence. Existing methods focus heavily on offline training. They aim to learn discriminative representations on the support set so that novel classes can be separated at test time. However, their discovery mechanism at inference is typically reduced to a single threshold. We argue that this paradigm is fundamentally flawed as OCD is not a static classification problem, but a dynamic process. The model must continuously decide 1) whether a sample belongs to a known class, 2) matches an existing novel category, or 3) should initiate a new one. Moreover, prior methods treat the support set as fixed knowledge. They do not update their decision boundaries as new evidence arrives during inference. This leads to unstable and inconsistent category formation. Our experiments confirm these issues. With properly calibrated and adaptive thresholds, substantial improvements can be achieved, even without changing the representation. Motivated by this, we propose \textbf{\method}, a support-set-calibrated, tree-structured online decision framework. The framework models inference as a sequence of hierarchical decisions, including known-class routing, birth-aware novel assignment, and attach-versus-create operations over a dynamic prototype memory. Furthermore, we simulate the proxy discovery process to initialize the thresholds during offline training to align with inference. Thresholds are continuously updated during inference using mature novel prototypes. Importantly, \method{} requires no heavy training and no dataset-specific tuning. It can be directly integrated into existing OCD pipelines as an inference-time module. Extensive experiments show significant improvements over SOTA baselines across seven benchmarks.
\end{abstract}

\maketitle

\section{Introduction}

Conventional visual recognition systems are built under the closed-world assumption, where training and test data share a fixed label space \cite{11,22,33,vaswani2017attention,lecun2015deep,herrera2016multilabel}. In dynamic environments, however, models inevitably encounter unlabeled concepts outside this space. A closed-set classifier can neither recognize such samples nor organize them into new semantic categories. Novel Category Discovery (NCD) \cite{NCD} and Generalized Category Discovery (GCD) \cite{2022GCD} address this by transferring knowledge from labeled known classes to unlabeled data. However, they assume access to a predefined query set and process it offline. This assumption is restrictive, as real-world data arrives as a stream and requires immediate decisions.

On-the-Fly Category Discovery (OCD) provides a more realistic formulation \cite{SMILE}. It consists of an offline stage, where the model learns from a labeled support set, and an online stage, where unlabeled samples arrive sequentially and may belong to either known or novel classes. Recent methods have made progress in this setting. SMILE uses hash-based category descriptors for instance-level prediction. PHE \cite{PHE} improves robustness with prototypical hash encoding, especially in fine-grained scenarios. DiffGRE \cite{DiffGRE} enhances offline learning by synthesizing pseudo-novel samples. AGE \cite{AGE} formulates OCD as a dual subproblem, enabling explicit disentanglement of
known and novel class inference.

Despite these advances, existing methods share a common limitation. They place most of the modeling effort on offline representation learning, while treating the online discovery process as a simple decision step. In practice, inference is often reduced to a single threshold that separates known and novel samples. This design overlooks the structure of OCD. Online discovery is not a one-shot decision, but a sequence of coupled decisions for each incoming sample. The model must determine 1) whether the sample should compete with known classes, 2) whether an existing novel cluster can explain it, and 3) when a new cluster should be created. Compressing these decisions into a single global matching rule makes the system inherently fragile, especially in fine-grained settings where inter-class gaps are small. As a result, even a slight bias in the decision boundary can absorb known-class samples into novel clusters. Errors made early in the stream are difficult to correct and tend to propagate over time, leading to fragmentation of known classes and proliferation of spurious novel clusters.

These observations suggest a different view of OCD. Learning a good feature space is only part of the problem. The other part is how to use that feature space during online inference. Once the representation is reasonably stable, properly calibrated decision boundaries can already improve discovery quality significantly, even without retraining the network or modifying model parameters. Based on this view, we propose \method, a support-set-calibrated, tree-structured online decision framework. Instead of relying on a single global rule, our method explicitly models inference as a sequence of hierarchical decisions. We first construct margin-aware spherical representations and derive a normalized coordinate system from support-set statistics. During online inference, each sample is processed through three stages: known-class evidence routing, birth-aware novel assignment, and attach-versus-create decisions over a dynamic prototype memory.

To initialize the decision thresholds, we simulate discovery via proxy tasks defined on the support set during the offline stage. This aligns the training procedure with the online inference process. In contrast to prior methods that rely on fixed knowledge learned offline, our framework updates decision thresholds during inference using mature novel prototypes. This allows the system to adapt as new evidence accumulates and leads to more stable category formation over time.
Our framework operates directly at the inference stage of existing OCD methods in a plug-and-play manner. It requires no heavy retraining, avoids dataset-specific hyperparameter tuning, and improves online decision quality at low cost. Experiments on seven OCD benchmarks demonstrate significant gains over representative baselines. These results suggest that, for open category discovery, explicit decision modeling and support-set calibration are as critical as representation learning. \textbf{Our contributions can be summarized as follows:}
\begin{itemize}
\item We revisit OCD from a decision-centric perspective and shift the focus from offline representation learning to online discovery mechanisms. We propose a unified calibration scheme for known-class routing, birth-aware novel assignment, and attach-versus-create decisions using proxy tasks constructed solely from the support set.
\item We introduce a training--inference alignment strategy by simulating discovery through proxy tasks during the offline stage, enabling principled initialization of decision thresholds.

\item We propose an online adaptation mechanism that updates decision thresholds using mature novel prototypes, allowing the system to evolve as new concepts are observed.

\item We demonstrate through extensive experiments that careful online decision modelling with calibrated thresholds provides a simple, efficient, and effective way to improve OCD performance.
\end{itemize}

\section{Related Work}
\subsection{On-the-Fly Category Discovery}
Novel Category Discovery (NCD) learns to organize unlabeled samples from novel classes by transferring supervision from labeled known classes \cite{NCD,NCD1,NCD2,NCD3}. Generalized Category Discovery (GCD) extends this setting by allowing the unlabeled pool to contain both known and novel classes \cite{2022GCD,zhao2023learning,rastegar2023learn,wu2023metagcd,wen2023parametric,vaze2023no,liu2025collaborative,shi2024unified}. Recent GCD research has expanded along several directions, including clustering-oriented representation learning \cite{choi2024contrastive}, efficient prompt-based adaptation \cite{wang2024sptnet,zhang2025less}, fine-grained and ultra-fine-grained discovery \cite{rastegar2024selex,liu2024novel}, data-centric or distributed settings \cite{pu2024federated,zhao2024labeled,cao2025allgcd}, alternative geometric priors \cite{liu2025hyperbolic}, language-guided or multi-modal transfer \cite{ouldnoughi2023clip,zheng2024textual,wang2025get}, scene-aware modeling \cite{peng2025mos}, and robustness to domain shift \cite{wang2024hilo,rathore2025domain}. Related open-world semi-supervised learning formulations also study novel-class discovery from partially labeled data \cite{cao2021open,rizve2022openldn}. Despite this methodological diversity, these formulations still assume a predefined unlabeled pool and perform discovery offline, which limits their suitability for dynamic environments where samples arrive sequentially, the label space evolves over time, and decisions must be made at the instance level \cite{ma1,ma2,ma3}.

To remove the dependence on a fixed unlabeled pool and support streaming inference, SMILE introduces On-the-Fly Category Discovery (OCD) \cite{SMILE}. Existing OCD methods mainly improve category descriptors or feature learning. SMILE uses hash codes as compact category descriptors for online assignment. PHE represents each category with multiple hash prototypes to reduce the instability of binary codes in fine-grained recognition. DiffGRE synthesizes pseudo-novel samples to strengthen offline training \cite{DiffGRE}. More recent methods further enrich this line from different angles. SynC incorporates language-assisted feature representation together with a lightweight active learning strategy, showing that semantic cues from language can improve online category discovery without introducing heavy annotation cost \cite{sync}. AGE reformulates OCD by explicitly disentangling known-class recognition and novel-class discovery, and models category densities with adaptive Gaussian expansion to support dynamic cluster growth during streaming inference \cite{AGE}. These advances improve the representation or category-modeling side of the problem, but they still leave the online decision layer relatively under-modeled. In contrast, we focus on how a model should route samples between known classes and evolving novel clusters during testing, and on how the corresponding decision boundaries can be calibrated from the support set.

\subsection{Threshold-based decisions}
Threshold-based decisions have long been used in open-set recognition, open-world learning\cite{tau1,tau2}, and streaming clustering to reject unknown samples, detect anomalies, or decide when to create new categories\cite{tau3,tau4}. Early open-set methods typically formulate rejection by thresholding an unknownness score or calibrated class confidence, as exemplified by OpenMax and related post-hoc decision rules\cite{tau5}. More recent distance- or prototype-based methods further improve the reliability of such decisions by explicitly shaping compact known-class regions in feature or logit space, so that rejection can be performed according to geometric compatibility with class anchors or prototypes\cite{miller2021class}. There is also growing evidence that threshold selection itself is highly task-dependent: in few-shot and open-world settings, different tasks may require very different rejection strengths, which makes manual tuning unstable and often impractical\cite{tau7,tau8}.

In most cases, thresholds are still treated as tunable hyperparameters, adjusted on validation data, estimated from assumptions about unknown distributions, or coupled with test-time updates. These assumptions do not fit OCD well, because before the stream starts the model typically has access only to the support set and cannot observe the true distribution of novel classes in advance. Our work differs in that it treats threshold calibration as a central part of online inference. We calibrate the thresholds for known-class routing, dynamic birth control, and attach-versus-create decisions through support-set proxy tasks, which yields a unified tree-structured decision process without manual threshold tuning.

\section{Method}
\label{sec:method}
\textbf{Problem definition.}
On-the-Fly Category Discovery (OCD) is formulated as a streaming recognition task where a model must maintain discriminative power over known classes while incrementally identifying novel categories from an unlabeled test stream. Formally, we are provided with a labeled support set $\mathcal{D}_S = \{(\mathbf{x}_i, y_i)\}_{i=1}^{N_S} \subseteq \mathcal{X} \times \mathcal{Y}_S$ for offline training. At inference, the model encounters an unlabeled query stream $\mathcal{D}_Q = \{(\mathbf{x}_j, y_j)\}_{j=1}^{N_Q} \subseteq \mathcal{X} \times \mathcal{Y}_Q$ in an online fashion, where $\mathcal{Y}_S \subset \mathcal{Y}_Q$. The label set $\mathcal{Y}_S$ denotes the base classes known a priori, while $\mathcal{Y}_{novel} = \mathcal{Y}_Q \setminus \mathcal{Y}_S$ represents novel categories that emerge only during deployment. Crucially, OCD requires \textit{immediate}, \textit{instance-level decisions}. Each sample in $\mathcal{D}_Q$ must be processed upon arrival without access to future data or a second pass. This temporal constraint distinguishes OCD from conventional closed-set recognition and offline discovery. For clarity, we refer to classes in $\mathcal{Y}_S$ as base classes in our methodology, while reserving the terms Old and New to denote the standard evaluation metrics.

\begin{figure*}[t]
  \centering
  \includegraphics[width=0.9\textwidth]{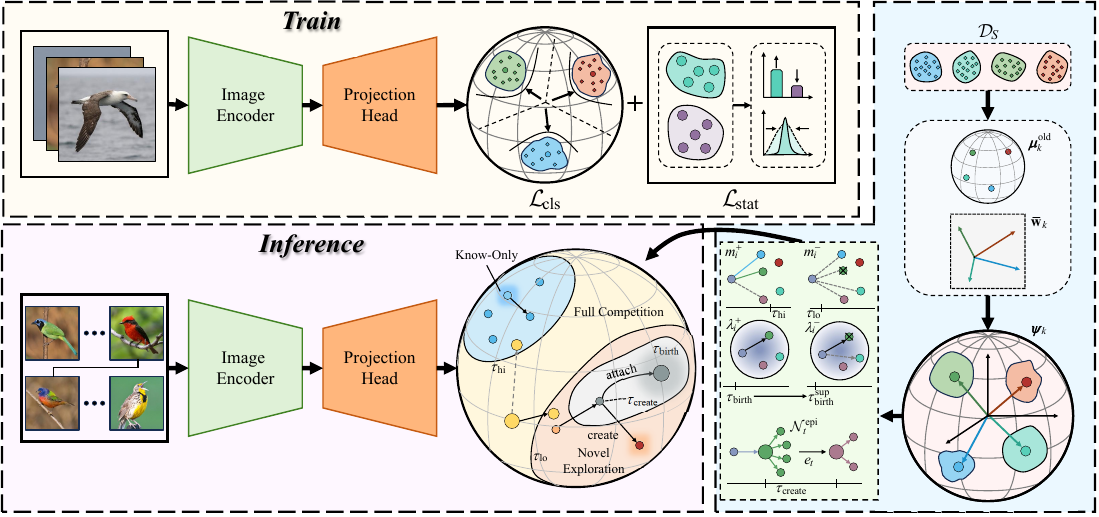}
  \vspace{-1em}
  \caption{Overview of the proposed framework. The model first learns a spherical representation from the support set. At test time, each streaming sample is routed by base-class evidence, checked by a support-calibrated birth boundary, and then either assigned to an existing prototype or used to create a new one. All key thresholds are estimated from support-only proxy tasks before online inference starts.}
  \Description{Framework overview with three parts: spherical fine-tuning on the support set, routing regions for old-only, global, and new-only decisions on the sphere, and online novel-cluster expansion with support-calibrated thresholds.}
  \label{fig:framework}
  \vspace{-1em}
\end{figure*}

\subsection{Standardized Spherical Representation}
\label{subsec:representation}
The effectiveness of category discovery hinges on a representation space that is both discriminative for known concepts and extensible for unseen ones. We represent all samples on a standardized spherical manifold, a choice motivated by two critical requirements of the OCD task. First, in an open-world stream, the magnitude of feature vectors often encapsulates instance-specific variance or prediction confidence rather than semantic identity. By constraining embeddings to a unit sphere, we decouple semantic direction from scale, ensuring that discovery depends purely on angular affinity. Second, spherical geometry, when optimized with angular margins, induces a compact intra-class structure that creates "vacant" angular regions. This structured sparsity is essential for accommodating novel prototypes without encroaching on the boundaries of base classes.

\textbf{Angular Margin Learning.} Given an input image $x$, we extract its visual feature $z = f_\theta(x) \in \mathbb{R}^{d_z}$ using a self-supervised pre-trained backbone and map it to a $d$-dimensional projected feature $h = g_\phi(z) \in \mathbb{R}^d$ via a linear projection head, where $d_z$ and $d$ denote the backbone feature dimension and the projected embedding dimension, respectively. To induce the desired geometry, we optimize the model on the support set $\mathcal{D}_S$ using an additive angular margin loss $\mathcal{L}_{\text{cls}}$ \cite{arc}. This objective minimizes the intra-class angular variance while maximizing the inter-class gap:
\begin{equation}
\mathcal{L}_{\text{cls}} = -\log \frac{\exp(s \cdot \cos(\theta_{c} + m))}{\exp(s \cdot \cos(\theta_{c} + m)) + \sum_{j \neq c} \exp(s \cdot \cos\theta_j)},
\label{eq:cls_loss}
\end{equation}
where $c$ denotes the ground-truth label of $x$, $\cos\theta_j = w_j^\top (h/\|h\|_2)$, and $s$ and $m$ denote the hyperspherical scale and margin, respectively. To further align the representation with discovery, we incorporate a statistical regularization $\mathcal{L}_{\text{stat}}$ (see Appendix), resulting in the total loss $\mathcal{L} = \mathcal{L}_{\text{cls}} + \lambda_{\text{stat}}\mathcal{L}_{\text{stat}}$.

\textbf{Coordinate Standardization.} While unit-norm normalization projects features onto a sphere, it is often insufficient to eliminate coordinate-wise scale biases inherited from self-supervised pre-trained backbones. Such biases can distort the angular distances and lead to suboptimal decision boundaries during online discovery. To establish an isotropic and robust metric basis, we standardize the embeddings using the global first- and second-order statistics of the support set $\mathcal{D}_S$. Specifically, we compute the support-set mean $\mathbf{m}_S$ and variance $\boldsymbol{\nu}_S$ as:
\begin{equation}
\mathbf{m}_S=\frac{1}{N_S}\sum_{i=1}^{N_S} h_i,\quad
\boldsymbol{\nu}_S=\frac{1}{N_S}\sum_{i=1}^{N_S}(h_i-\mathbf{m}_S)\odot(h_i-\mathbf{m}_S),
\label{eq:support_stats}
\end{equation}
where $\odot$ denotes element-wise multiplication. Any projected feature $h$ is then mapped to the standardized spherical space, yielding the final spherical embedding $u$ as:
\begin{equation}
\tilde h=\big(h-\mathbf{m}_S\big)\odot \boldsymbol{\nu}_S^{-1/2},
\quad
u=\frac{\tilde h}{\|\tilde h\|_2}.
\label{eq:std_embed}
\end{equation}

This standardization procedure ensures that all subsequent operations operate within a unified, zero-centered, and variance-calibrated spherical manifold, thereby enabling robust hierarchical routing and dynamic prototype creation. Such alignment is essential for the plug-and-play nature of our framework, as it harmonizes the query stream's distribution with the metric structure of the support set.

\subsection{Hierarchical Online Decision}
\label{subsec:inference}
Building upon the standardized spherical manifold, we reformulate the inference process as a hierarchical decision tree rather than a static, flat classification. This approach acknowledges that OCD is a non-stationary process where the model must continuously adjudicate between retaining known knowledge and accommodating emerging concepts. We maintain a dynamic prototype memory $\mathcal{M}_t$ at time step $t$ consisting of $K_{\text{base}}$ fixed base-class prototypes and $K_{\text{novel}}^{(t)}$ evolving novel prototypes.

At each time step $t$, the incoming sample is represented by its standardized spherical embedding $u_t$ as derived through the standardization procedure in Sec.~\ref{subsec:representation}. To mitigate decision instability caused by local geometric noise and temporal fluctuations in the stream, we augment the cosine similarity $\ell_k(u_t)$ between $u_t$ and prototype $k$ with a Dirichlet-smoothed size prior \cite{li2019tutorial}. For each prototype $k$ within the dynamic memory bank $\mathcal{M}_t$, the comprehensive matching score $s_k(u_t)$ is formulated as:
\begin{equation}
\pi_k(t) = \frac{n_k(t) + \alpha}{\sum_{j=1}^{K_t} n_j(t) + K_t \alpha},\quad
s_k(u_t) = \frac{\ell_k(u_t)}{T} + \log \pi_k(t),
\label{eq:memory_score}
\end{equation}
where each prototype together with the samples currently assigned to it is treated as a cluster, $n_k(t)$ is the size of that cluster,  $T$ and $\alpha$ denote the temperature scaling parameter and the smoothing constant, respectively. This score combines local geometric similarity with global memory information, which makes online competition more stable.

\textbf{Evidence-Based Candidate Routing.}
In fine-grained scenarios, permitting unstable novel prototypes to compete globally with well-established base classes often leads to semantic leakage and the fragmentation of base categories. We mitigate this risk through a routing gate that restricts the candidate prototype set $C_t$ using base-class evidence. We compute the maximum base-class similarity $g_{\text{cos}}(u_t) = \max_{k \le K_{\text{base}}} \ell_k(u_t)$ and the top-1-to-2 margin $g_{\text{mar}}(u_t) = g_{\text{cos}}^{(1)}(u_t) - g_{\text{cos}}^{(2)}(u_t)$. The routing logic is orchestrated as follows:
\vspace{-0.5em}
\begin{equation}
C_t=
\begin{cases}
\{1,\dots,K_{\text{base}}\}, & g_{\mathrm{mar}}(u_t)\ge \tau_{\text{hi}},\\[4pt]
\{K_{\text{base}}+1,\dots,K_t\}, & g_{\mathrm{cos}}(u_t)< \tau_{\text{lo}},\\[4pt]
\{1,\dots,K_t\}, & \text{otherwise}.
\end{cases}
\label{eq:candidate_route}
\end{equation}

Within this hierarchy, $\tau_{\text{hi}}$ represents a high-confidence threshold for base-class membership, whereby a decisive margin ensures the sample is processed exclusively by known prototypes. Conversely, $\tau_{\text{lo}}$ acts as a conservative lower bound for discovered known-class affinity, ensuring that samples with negligible similarity to the support set are diverted directly to the discovery branch. By shielding the inference process from unnecessary global competition, this design safeguards confident base assignments and invokes full competition only when the evidence is ambiguous.

\textbf{Birth-Aware Novel Assignment.} For the second and third conditions in $C_t$, the framework determines whether the current prototype memory sufficiently explains the sample or if a birth event (creating a new class) is warranted. We formulate a birth statistic $\Lambda(u_t)$ grounded in a von Mises-Fisher (vMF) probabilistic interpretation \cite{banerjee2005clustering}:
\begin{equation}
\Lambda(u_t) = \max_{k \in C_t} \frac{\ell_k(u_t)}{T} - \log p_0,
\label{eq:birth_stat}
\end{equation}
where $p_0$ represents the uniform background density on the unit sphere; its closed-form computation and vMF interpretation are given in Appendix~\ref{app:vmf}. If $\Lambda(u_t) \ge \tau_{\text{birth}}^{(t)}$, the sample is assigned to the best-matching prototype $\hat y_t = \arg\max_{k \in C_t} s_k(u_t)$. Otherwise, it is diverted to the discovery regime as a potential pioneer of a new category.

\textbf{Attach-versus-Create Operation.} Within the discovery regime, the model evaluates whether the sample aligns with an existing novel category or justifies the creation of a new one. If no novel prototype exists yet, that is, $K_t = K_{\text{base}}$, we directly create a new one. Otherwise, let the active novel set be $\mathcal{N}_t=\{K_{\text{base}}+1,\dots,K_t\}$. For each $k\in\mathcal{N}_t$, we maintain its cluster size $n_k(t)$ and running resultant vector:
\begin{equation}
R_k(t)=\sum_{\tau < t:\hat y_\tau = k} u_\tau,
\qquad
\mu_k(t)=\frac{R_k(t)}{\|R_k(t)\|_2}.
\label{eq:novel_resultant}
\end{equation}
The cluster concentration $\kappa_k(t)$ is estimated from $R_k(t)$ and $n_k(t)$; the full expression is given in Appendix~\ref{app:attach}. We then define the concentration-aware matching score
\begin{equation}
a_k(u_t) = \log n_k(t) + \kappa_k(t) \mu_k^\top u_t - \log p_0.
\label{eq:attach_score}
\end{equation}
Here $p_0$ is the same background term as in Appendix~\ref{app:vmf}. We further define
\begin{equation}
k_t^\dagger=\arg\max_{k\in\mathcal{N}_t} a_k(u_t).
\label{eq:attach_argmax}
\end{equation}
We attach the sample to prototype $k_t^\dagger$ when $a_{k_t^\dagger}(u_t)\ge \tau_{\text{create}}$; otherwise we create a new prototype:
\begin{equation}
\hat y_t=
\begin{cases}
k_t^\dagger, & a_{k_t^\dagger}(u_t)\ge \tau_{\text{create}},\\[4pt]
K_t+1, & \text{otherwise}.
\end{cases}
\label{eq:attach_decision}
\end{equation}
This rule lets a sample enter the discovery branch without forcing immediate prototype creation.

\textbf{Online Memory Evolution.} Upon determining the assignment $\hat y_t$, we perform a single-pass update to reflect the new evidence. If $\hat y_t \le K_{\text{base}}$, we only update its visit count and keep the base prototype fixed. If $\hat y_t$ corresponds to an existing novel prototype, we update its memory by
\begin{equation}
n_{\hat y_t}(t+1)=n_{\hat y_t}(t)+1,
\qquad
R_{\hat y_t}(t+1)=R_{\hat y_t}(t)+u_t,
\label{eq:update_existing_nR}
\end{equation}
\begin{equation}
\mu_{\hat y_t}(t+1)=
\frac{R_{\hat y_t}(t+1)}{\|R_{\hat y_t}(t+1)\|_2}.
\label{eq:update_existing_mu}
\end{equation}
If the current sample creates a new prototype, we initialize
\begin{equation}
n_{K_t+1}(t+1)=1,\qquad
R_{K_t+1}(t+1)=u_t,\qquad
\mu_{K_t+1}(t+1)=u_t.
\label{eq:init_new_proto}
\end{equation}

This update mechanism allows the semantic centroids and concentration estimates of novel categories to evolve synchronously with the streaming data, enabling the model to adapt its decision boundaries as concepts mature.

\subsection{Principled Calibration via Proxy Tasks}
\label{subsec:calibration}

A pivotal advantage of our framework is the automated initialization of decision thresholds without dataset-specific hyperparameter tuning. We achieve this by constructing Proxy Discovery Tasks on the labeled support set $\mathcal{D}_S$. These tasks simulate the online discovery process by treating known data as a surrogate for unseen concepts, thereby grounding the thresholds in the empirical density of the learned representation.

For any decision statistic $q(\cdot)$, we derive the optimal threshold $\tau_q$ by maximizing the balanced accuracy over a synthetic response set. In this optimization, $\tau$ represents a candidate decision boundary defined within the scalar range of the statistic $q$. Let $P_q$ and $N_q$ be the sets of positive and negative responses generated from the support set, respectively. The threshold is selected as:
\begin{equation}
\tau_q = \arg\max_\tau \frac{1}{2} \left( \frac{1}{|P_q|} \sum_{r \in P_q} \mathbb{I}[r \ge \tau] + \frac{1}{|N_q|} \sum_{r' \in N_q} \mathbb{I}[r' < \tau] \right).
\label{tau_q_compu}
\end{equation}

This criterion ensures that the resulting decision boundaries maximize the separation between correct assignments and likely discovery errors. By grounding the selection in the empirical distribution of the support set, this procedure provides a principled initialization that is inherently aligned with the density of the representation space.

\textbf{Routing Calibration ($\tau_{\text{hi}}, \tau_{\text{lo}}$).} 
To calibrate the routing gate, we employ a leave-one-out strategy that emulates class absence within the support set $\mathcal{D}_S$. For each support sample $(u_i, c_i)$, we first compute the positive margin response $m_i^{+} = g_{\text{cos},i}^{(1)} - g_{\text{cos},i}^{(2)}$ by including the ground-truth prototype in the calculation. To simulate the occurrence of a novel category, we mask the true class $c_i$ and recompute the top-1 and top-2 similarities among the remaining classes to obtain the negative margin response $m_i^{-} = g_{\text{cos},i, \setminus c_i}^{(1)} - g_{\text{cos},i, \setminus c_i}^{(2)}$.

In this context, $\{m_i^{+}\}$ represents the margins associated with confident base-class evidence, whereas $\{m_i^{-}\}$ captures the margin distribution when the model is forced to explain a sample via the most similar distractors. We derive the high-confidence threshold $\tau_{\text{hi}}$ by maximizing the balanced accuracy $\mathcal{B}_{\mathrm{mar}}(\tau)$ as follows:
\begin{equation}
\mathcal{B}_{\mathrm{mar}}(\tau) = \frac{1}{2} \left[ \frac{1}{N_S} \sum_{i=1}^{N_S} \mathbb{I}[m_i^{+} \ge \tau] + \frac{1}{N_S} \sum_{i=1}^{N_S} \mathbb{I}[m_i^{-} < \tau] \right],
\end{equation}

This optimization identifies the optimal decision boundary that distinguishes legitimate base assignments from cases where the base-side evidence is merely a byproduct of class absence. While $\tau_{\text{hi}}$ serves as a discriminative separator, the routing gate also requires a floor threshold $\tau_{\text{lo}}$ to identify samples clearly unsupported by any known class. Unlike the symmetric nature of $\tau_{\text{hi}}$, $\tau_{\text{lo}}$ is designed as a conservative lower envelope of base-class affinity. Let $r_i^{+} = \max_{k \le K_{\text{base}}} \ell_k(u_i)$ denote the maximum base-class similarity for each support sample. We then formulate $\tau_{\text{lo}}$ as:
\begin{equation}
\tau_{\text{lo}}
=
\min\!\left\{
\tau_{\text{hi}},
\ \min_i r_i^{+}-\mathrm{Std}_i[r_i^{+}]
\right\}.
\end{equation}

This rule ensures that a sample is diverted to the discovery branch only when its maximum similarity to all known prototypes falls significantly below the lower bound of normal support-set responses. Through this two-tiered calibration, the candidate routing in Sec.~\ref{subsec:inference} is determined automatically from the statistics of $\mathcal{D}_S$ without the need for manual parameter search.

\begin{table*}[!t]
  \centering
  {\small
  \setlength{\tabcolsep}{6pt}
  \renewcommand{\arraystretch}{1.25}
    \caption{Comparison with other SOTA methods under the Greedy--Hungarian and Strict--Hungarian protocols.
    ``DiffGRE-S'' integrates DiffGRE with SMILE, while ``DiffGRE-P'' integrates DiffGRE with PHE.
    Average denotes the mean across all seven datasets; missing entries are marked with ``--''.
    The \best{best} results are highlighted in red, and the \second{second-best} results in blue.
    \underline{Underlined numbers} indicate results not reported in the original papers but reproduced using the official code.}
  \label{tab:sota_comparison}
  \vspace{-1em}
  \resizebox{\textwidth}{!}{
  \begin{tabular}{cc
  |ccc|ccc|ccc|ccc|ccc|ccc|ccc|ccc}
  \toprule
  \multirow{2}{*}{}
  & \multirow{2}{*}[-0.6ex]{\textbf{Method}}
  & \multicolumn{3}{c|}{\textbf{CIFAR-10 (\%)}}
  & \multicolumn{3}{c|}{\textbf{CIFAR-100 (\%)}}
  & \multicolumn{3}{c|}{\textbf{ImageNet-100 (\%)}}
  & \multicolumn{3}{c|}{\textbf{CUB-200-2011 (\%)}}
  & \multicolumn{3}{c|}{\textbf{Stanford Cars (\%)}}
  & \multicolumn{3}{c|}{\textbf{Oxford Pets (\%)}}
  & \multicolumn{3}{c|}{\textbf{Food101 (\%)}}
  & \multicolumn{3}{c}{\textbf{Average (\%)}} \\
  \cmidrule(lr){3-5}\cmidrule(lr){6-8}\cmidrule(lr){9-11}\cmidrule(lr){12-14}\cmidrule(lr){15-17}\cmidrule(lr){18-20}\cmidrule(lr){21-23}\cmidrule(lr){24-26}
  &
  & All & Old & New
  & All & Old & New
  & All & Old & New
  & All & Old & New
  & All & Old & New
  & All & Old & New
  & All & Old & New
  & All & Old & New \\
  \midrule
  \multirow{6}{*}{\rotatebox[origin=c]{90}{\textit{Greedy--Hungarian}}}
  & SLC \cite{hartigan1975clustering}
  & 65.9 & 96.5 & 50.9
  & 46.9 & 62.1 & 16.6
  & 34.2 & 86.6 & 7.1
  & 31.3 & 48.5 & 22.7
  & 24.0 & 45.8 & 13.6
  & -- & -- & --
  & -- & -- & --
  & -- & -- & -- \\
  & MLDG \cite{MLDGli2018learning}
  & 71.6 & \second{97.5} & 58.6
  & 58.4 & 69.0 & 37.3
  & 33.6 & 74.4 & 13.1
  & 34.2 & 57.9 & 22.4
  & 28.0 & 49.1 & 17.7
  & -- & -- & --
  & -- & -- & --
  & -- & -- & -- \\
  & RankStat \cite{RankStat}
  & 56.5 & 81.1 & 44.2
  & 36.9 & 45.7 & 19.3
  & 33.1 & 74.2 & 12.4
  & 27.6 & 46.2 & 18.3
  & 18.6 & 36.9 & 9.7
  & -- & -- & --
  & -- & -- & --
  & -- & -- & -- \\
  & WTA \cite{WTAjia2021joint}
  & 65.4 & 88.0 & 54.1
  & 44.1 & 55.5 & 21.2
  & 33.1 & 75.8 & 11.7
  & 26.5 & 45.0 & 17.3
  & 20.0 & 38.8 & 10.6
  & -- & -- & --
  & -- & -- & --
  & -- & -- & -- \\
  & SMILE \cite{SMILE}
  & \second{78.2} & \best{99.3} & \second{67.6}
  & \second{61.3} & \second{70.7} & \second{42.5}
  & \second{39.9} & \second{87.1} & \second{16.2}
  & \second{41.1} & \second{67.6} & \second{27.8}
  & \second{33.4} & \second{58.4} & \second{21.3}
  & \second{54.1} & \second{66.1} & \second{47.8}
  & \second{34.4} & \second{64.0} & \second{19.4}
  & \second{48.9} & \second{73.3} & \second{34.7} \\
  & \textbf{\method}
  &\cellcolor{ltgray}\best{87.1} & \cellcolor{ltgray} 97.4 & \cellcolor{ltgray}\best{82.0}
  & \cellcolor{ltgray}\best{71.1} & \cellcolor{ltgray}\best{83.6} & \cellcolor{ltgray}\best{46.1}
  & \cellcolor{ltgray}\best{68.9} & \cellcolor{ltgray}\best{92.7} & \cellcolor{ltgray}\best{56.9}
  & \cellcolor{ltgray}\best{53.8} & \cellcolor{ltgray}\best{79.7} & \cellcolor{ltgray}\best{40.9}
  & \cellcolor{ltgray}\best{53.1} & \cellcolor{ltgray}\best{79.1} & \cellcolor{ltgray}\best{40.5}
  & \cellcolor{ltgray}\best{78.1} & \cellcolor{ltgray}\best{85.7} & \cellcolor{ltgray}\best{74.2}
  & \cellcolor{ltgray}\best{44.0} & \cellcolor{ltgray}\best{72.7} & \cellcolor{ltgray}\best{29.3}
  & \cellcolor{ltgray}\best{65.1} & \cellcolor{ltgray}\best{84.4} & \cellcolor{ltgray}\best{52.8} \\
  \midrule
  \multirow{11}{*}{\rotatebox[origin=c]{90}{\textit{Strict--Hungarian}}}
  & SLC \cite{hartigan1975clustering}
  & 41.5 & \best{58.3} & 33.3
  & 44.4 & 59.0 & 15.1
  & 32.9 & 86.6 & 5.2
  & 28.6 & 44.0 & 20.9
  & 14.0 & 23.0 & 9.7
  & 35.5 & 41.3 & 33.1
  & 20.9 & 48.6 & 6.8
  & 31.1 & 51.5 & 17.7 \\
  & MLDG \cite{MLDGli2018learning}
  & 44.1 & 38.5 & 47.0
  & 50.6 & 61.0 & 29.8
  & 30.6 & 72.3 & 9.7
  & 29.5 & 48.4 & 20.1
  & 24.0 & 41.6 & 15.4
  & -- & -- & --
  & -- & -- & --
  & -- & -- & -- \\
  & RankStat \cite{RankStat}
  & 42.1 & 49.3 & 38.6
  & 35.0 & 44.0 & 17.0
  & 31.1 & 73.3 & 9.8
  & 21.2 & 26.9 & 18.4
  & 14.8 & 19.9 & 12.3
  & 33.2 & 42.3 & 28.4
  & 22.3 & 50.7 & 7.8
  & 28.5 & 43.8 & 18.9 \\
  & WTA \cite{WTAjia2021joint}
  & 43.1 & 34.5 & 47.4
  & 40.8 & 52.9 & 16.7
  & 30.8 & 72.9 & 9.7
  & 21.9 & 26.9 & 19.4
  & 17.1 & 24.4 & 13.6
  & 35.2 & 46.3 & 29.3
  & 18.2 & 40.5 & 6.1
  & 29.6 & 42.6 & 20.3 \\
  & SMILE \cite{SMILE}
  & \underline{49.9} & \underline{39.9} & \underline{54.9}
  & 51.6 & 61.6 & \second{31.7}
  & 33.8 & 74.2 & 13.5
  & 32.2 & 50.9 & 22.9
  & 26.2 & 46.6 & 16.3
  & 42.9 & 38.7 & 45.1
  & \underline{24.2} & \underline{54.3} & \underline{8.8}
  & 37.3 & 52.3 & 27.6 \\
  & PHE \cite{PHE}
  & \second{\underline{53.1}} & \underline{19.3} & \best{\underline{70.0}}
  & \underline{56.0} & \underline{70.1} & \underline{27.8}
  & \underline{39.2} & \underline{49.3} & \second{\underline{34.1}}
  & 36.4 & 55.8 & 27.0
  & 31.3 & 61.9 & 16.8
  & 48.3 & 53.8 & 45.4
  & 29.1 & 64.7 & 11.1
  & 41.9 & 53.6 & 33.2 \\
  & DiffGRE-S \cite{DiffGRE}
  & -- & -- & --
  & -- & -- & --
  & -- & -- & --
  & 35.4 & 58.2 & 23.8
  & 30.5 & 59.3 & 16.5
  & 42.4 & 42.1 & 42.5
  & -- & -- & --
  & -- & -- & -- \\
  & DiffGRE-P \cite{DiffGRE}
  & -- & -- & --
  & -- & -- & --
  & -- & -- & --
  & 37.9 & 57.0 & 28.3
  & 32.1 & \second{63.3} & 16.9
  & 48.6 & 52.6 & 46.6
  & -- & -- & --
  & -- & -- & -- \\
  & SynC \cite{sync}
  & 50.1 & 40.6 & 54.9
  & 56.1 & 68.4 & 31.5
  & 44.0 & 86.2 & 22.8
  & 45.3 & 54.3 & \second{40.9}
  & 24.6 & 34.8 & 19.6
  & \second{61.7} & \best{69.5} & 57.6
  & \second{31.1} & 60.7 & \second{16.0}
  & 44.7 & 59.2 & \second{34.8} \\
  & AGE \cite{AGE}
  & \underline{39.2} & \underline{17.2} & \underline{50.3}
  & \second{60.8} & \second{75.8} & 30.7
  & \second{48.2} & \second{87.1} & 28.6
  & \second{46.3} & \second{59.8} & 39.4
  & \second{34.8} & 62.7 & \second{21.3}
  & 61.5 & 66.5 & \second{58.8}
  & 30.5 & \best{70.0} & 10.4
  & \second{45.9} & \second{62.7} & 34.2 \\
  & \textbf{\method}
  & \cellcolor{ltgray}\best{58.0} & \cellcolor{ltgray}\second{57.6} & \cellcolor{ltgray}\second{58.3}
  & \cellcolor{ltgray}\best{67.0} & \cellcolor{ltgray}\best{78.8} & \cellcolor{ltgray}\best{43.5}
  & \cellcolor{ltgray}\best{67.8} & \cellcolor{ltgray}\best{92.7} & \cellcolor{ltgray}\best{55.3}
  & \cellcolor{ltgray}\best{49.7} & \cellcolor{ltgray}\best{67.1} & \cellcolor{ltgray}\best{41.0}
  & \cellcolor{ltgray}\best{49.9} & \cellcolor{ltgray}\best{72.4} & \cellcolor{ltgray}\best{39.0}
  & \cellcolor{ltgray}\best{71.8} & \cellcolor{ltgray}\second{67.4} & \cellcolor{ltgray}\best{74.2}
  & \cellcolor{ltgray}\best{40.0} & \cellcolor{ltgray}\second{69.0} & \cellcolor{ltgray}\best{25.2}
  & \cellcolor{ltgray}\best{57.7} & \cellcolor{ltgray}\best{72.1} & \cellcolor{ltgray}\best{48.1} \\
  \bottomrule
  \end{tabular}
  }
  }
\end{table*}

\textbf{Discovery and Creation Calibration ($\tau_{\text{birth}}, \tau_{\text{create}}$).} We initialize the thresholds for the discovery regime by simulating the transition from known classes to newly formed clusters using the support set $\mathcal{D}_S$. To calibrate the birth threshold $\tau_{\text{birth}}$, we evaluate the model's confidence in explaining a sample when its ground-truth category is excluded from the candidate set. For each support sample $u_i$, we define the positive and negative birth statistics as: 
\begin{equation}
\lambda_i^{+}=\max_{1\le k\le K_{\text{base}}}\frac{\ell_k(u_i)}{T}-\log p_0,
\qquad
\lambda_i^{-}=\max_{k\neq c_i}\frac{\ell_k(u_i)}{T}-\log p_0,
\end{equation}
where $\lambda_i^{+}$ represents the baseline compatibility when the true concept is available, whereas $\lambda_i^{-}$ measures the residual affinity when the sample is treated as a novel concept. We obtain the static support-based birth threshold $\tau_{\text{birth}}^{\text{sup}}$ by designating $\{\lambda_i^{+}\}$ and $\{\lambda_i^{-}\}$ as the positive and negative response sets, respectively, and optimizing the balanced accuracy $\mathcal{B}_{\text{birth}}(\tau)$ according to Eq.~\eqref{tau_q_compu}.

To calibrate the creation threshold $\tau_{\text{create}}$, we perform Pseudo-Novel Replays by treating each known class in $\mathcal{D}_S$ as a temporary novel category. We simulate an online stream by replaying support samples in random order and maintaining an episodic novel memory $\mathcal{N}_t^{\text{epi}}$. At each replay step $t$, we compute the best attach score $e_t$ against the current episodic prototypes:
\begin{equation}
e_t = \max_{k \in \mathcal{N}_t^{\text{epi}}} a_k(u_t).
\end{equation}

We collect the scores obtained during the first encounter of each class to form the negative response set $N_{\text{create}}$ because these samples should ideally trigger a creation event. Conversely, scores from subsequent encounters of the same class form the positive response set $P_{\text{create}}$ because they should be attached to the existing episodic prototype. The threshold $\tau_{\text{create}}$ is then derived by maximizing the balanced accuracy $\mathcal{B}_{\text{create}}(\tau)$ over these sets, which enables the model to distinguish between expanding an established cluster and initiating a new one. This simulation strategy ensures that all key parameters in the discovery tree are grounded in the empirical density of the support set, thereby facilitating a principled transition from offline training to online inference.

\subsection{Adaptive Decision Boundary}
\label{subsec:adaptive_boundary}

While support-set calibration provides a robust prior, the online stream is inherently non-stationary. To maintain stability, \method{} adaptively refines the birth boundary $\tau_{\text{birth}}^{(t)}$ as novel prototypes mature. We consider a novel prototype $k$ to be mature if its cardinality exceeds a stability threshold: $n_k(t) \ge \text{round} ( (n_{\text{base}}^{\text{med}})^\beta )$, where $n_{\text{base}}^{\text{med}}$ is the median support size of base classes. For each mature prototype, we compute its self-explanation strength $\lambda_k^{\text{self}}(t)$ as:
\begin{equation}
\lambda_k^{\text{self}}(t)=
\frac{n_k(t)-1}{n_k(t)+1}
\cdot
\frac{\|R_k(t)\|_2}{n_k(t)}
\cdot
\frac{1}{T}
-\log p_0.
\label{eq:self_explain_strength}
\end{equation}

This metric describes the internal semantic cohesion of the discovered cluster. We aggregate these statistics into a bank reference $\tau_{\text{birth}}^{\text{bank}}(t)$ and update the effective threshold using a momentum-based blend:
\begin{equation}
\tau_{\text{birth}}^{(t)}=
\min\!\left\{
\tau_{\text{birth}}^{\text{sup}},
\ (1-\eta_t)\tau_{\text{birth}}^{\text{sup}}+\eta_t\tau_{\text{birth}}^{\text{bank}}(t)
\right\}.
\label{eq:dynamic_birth_threshold}
\end{equation}

The trust factor $\eta_t$ scales with the relative maturity of the novel bank, which allows the model to progressively transition from the support-set prior to the dynamic statistics of the online stream. This adaptation suppresses spurious births while ensuring that the decision boundaries remain aligned with the evolving category structure.

\section{Experiments}
\label{sec:experiments}
\subsection{Datasets and Setup}
\label{subsec:exp_setup}
\textbf{Datasets.}
We evaluate our method on seven standard OCD benchmarks, including three coarse-grained datasets, CIFAR-10, CIFAR-100\cite{cifar}, and ImageNet-100\cite{2015imagenet}, and four fine-grained datasets, CUB-200-2011\cite{CUB}, Stanford Cars\cite{scars}, Oxford-IIIT Pet\cite{pets}, and Food-101\cite{food}. Following the conventional OCD protocol \cite{SMILE,DiffGRE,PHE}, each dataset is split into known and novel categories. For the known categories, $50\%$ of the training samples are used to form the labeled support set $\mathcal{D}_S$, while the remaining samples together with all novel-category samples form the unlabeled query stream $\mathcal{D}_Q$. The stream therefore contains a mixture of old and new classes and is processed sequentially in a strict single-pass manner.

\textbf{Evaluation Metrics.}
Following  \cite{SMILE}, we adopt two protocols for evaluation termed Greedy-Hungarian and Strict-Hungarian for comprehensive comparisons, where their difference is clearly illustrated in \cite{2022GCD}. During testing, samples sharing the same category descriptor form a cluster, and only the top-$|Y_Q|$ clusters by size are retained, with the rest treated as misclassified. For Greedy-Hungarian, accuracy is computed separately on the "Old" and "New" subsets, providing independent evaluations of known and novel classes. In contrast, Strict-Hungarian calculates accuracy over the entire query set, avoiding repeated cluster assignments between subsets. The overall accuracy is obtained via the Hungarian matching:
\begin{equation}
ACC = \max_{p \in \mathcal{P}(\mathcal{Y}Q)} \frac{1}{|\mathcal{D}Q|} \sum_{i=1}^{|\mathcal{D}Q|} \mathbb{I}[y_i = p(\hat y_i)],
\end{equation}
where $y_i$ is the ground truth label, $\hat{y}_i$ is the predicted label decided by cluster indices, and $P(Y_Q)$ is the set of all permutations of ground truth labels.

\textbf{Implementation Details.}
 Unless stated otherwise, all experiments utilize a Vision Transformer backbone (ViT-B/16) pre-trained with DINO \cite{DINO}, featuring a $768$-dimensional linear projection head. The angular-margin classifier is trained with a hyperspherical scale $s = 30.0$ and a margin $m = 0.5$. During online inference, we maintain a consistent hyperparameter configuration across all seven datasets: the score temperature is fixed to $T = 1.0$, the Dirichlet smoothing constant is set to $\alpha = 10^6$, and the maturity exponent is defined as $\beta = 0.5$. The pseudo-novel replay mechanism for calibrating $\tau_{\text{create}}$ consists of three simulated passes over the support set. Importantly, all known-class directions and decision thresholds remain fixed once support calibration is completed, ensuring that our framework operates in a truly tuning-free and automated manner across diverse semantic domains.

\vspace{-0.5em}
\subsection{Comparison with State-of-the-Art}

\begin{table}[tbp]
\small
\centering
\setlength{\tabcolsep}{2pt}
\caption{Comparison of estimated category numbers and accuracies. Prior OCD methods exhibit "category explosion" with SMILE-32bit predicting 2953 clusters for Stanford Cars ($C=196$). PACO provides more realistic estimates including 194 for CUB ($C=200$). This stability results from dynamic birth control and threshold calibration which effectively suppress redundant prototype instantiation and ensure high fidelity to the true class distribution.}
  \vspace{-0.5em}
\label{tab:pred_acc_count}
\resizebox{0.45\textwidth}{!}{
\begin{tabular}{l|c|ccc|c|ccc}
\toprule
\multirow{2}{*}[-0.6ex]{Method}
& \multicolumn{4}{c|}{CUB ($C{=}200$)}
& \multicolumn{4}{c}{SCars ($C{=}196$)} \\
\cmidrule(lr){2-5} \cmidrule(lr){6-9}
& \#Cls & All & Old & New & \#Cls & All & Old & New \\
\midrule
SMILE-16bit~\cite{SMILE}    
& 924 & 31.9 & 52.7 & 21.5 & 896 & 27.5 & 52.5 & 15.4 \\
SMILE-32bit~\cite{SMILE}    
& 2146 & 27.3 & 52.0 & 15.0 & 2953 & 21.9 & 46.8 & 9.9 \\
PHE-16bit~\cite{PHE}        
& 318 & 37.6 & 57.4 & 27.6 & 709 & 31.8 & 65.4 & 15.6 \\
PHE-32bit~\cite{PHE}        
& 474 & 38.5 & 59.9 & 27.8 & 762 & 31.5 & 64.0 & 15.8 \\
SynC~\cite{sync}
& 444 & 45.3 & 54.3 & 40.9 & 484 & 24.6 & 34.8 & 19.6 \\
AGE~\cite{AGE}
& 397 & 46.3 & 59.8 & 39.4 & 633 & 34.8 & 62.7 & 21.3 \\
\rowcolor{gray!10}
\textbf{\method}
& \textbf{194} & \textbf{49.7} & \textbf{67.1} & \textbf{41.0}
& \textbf{367} & \textbf{49.9} & \textbf{72.4} & \textbf{39.0} \\
\bottomrule
\end{tabular}
}
  \vspace{-1em}
\end{table}

\textbf{Overall Performance.} As summarized in Table~\ref{tab:sota_comparison}, \method{ } consistently delivers superior performance across all seven benchmark datasets, establishing a new state-of-the-art for OCD task. Under the rigorous Strict-Hungarian protocol, our framework achieves the highest average results of 57.7\% in All, 72.1\% in Old, and 48.1\% in New accuracy, significantly outperforming competitive baselines such as AGE (45.9\% All) and SynC (44.7\% All). The substantial improvement in New accuracy (averaging 48.1\%) is particularly insightful, as it demonstrates that our calibrated decision logic effectively identifies emerging categories without relying on heuristic quantization, which often diminishes representational expressiveness. This trend remains robust in fine-grained scenarios like CUB-200-2011, where \method{} achieves the highest New-class accuracy of 41.0\%, successfully navigating the narrow inter-class boundaries and label space shifts inherent in non-stationary data streams.

\textbf{Category-Number Estimation.} 
To further evaluate the quality of discovery, we analyze the estimated category numbers ($\#Cls$) in Table~\ref{tab:pred_acc_count}, which highlights a notorious limitation of prior OCD frameworks known as category explosion. Hash-based methods like SMILE and PHE exhibit severe over-segmentation; for instance, on Stanford Cars ($C=196$), SMILE-32bit predicts a staggering 2953 clusters, while PHE-32bit still fragments the space into 762 pseudo-classes. In contrast, \method{} provides a significantly more stable and realistic estimate, predicting 194 categories on CUB (remarkably close to the ground truth of $C=200$) and 367 on Stanford Cars. This stability is primarily attributed to our dynamic birth control and threshold calibration, which successfully suppress redundant prototype instantiation. By ensuring each created prototype aligns with a genuine semantic concept, \method{} achieves a much higher fidelity to the true class distribution than recent baselines like SynC or AGE, proving that precise inference-time boundaries are vital for successful open-world learning.

\begin{table}[tbp]
\small
\centering
\setlength{\tabcolsep}{3pt}
\caption{Ablation study on CUB and Stanford Cars. The results validate the contribution of each module, with the angular-margin objective being most critical for establishing spherical geometry. Statistical regularization and OldRefSel stabilize Old-class accuracy, while EvidenceRouting and DynBirthTight optimize New-class discovery and mitigate redundant prototype instantiation.}
\label{tab:ablation_cub_scars}
  \vspace{-0.5em}
\resizebox{0.45\textwidth}{!}{
\begin{tabular}{l|ccc|ccc}
\toprule
\multirow{2}{*}[-0.6ex]{Variant}
& \multicolumn{3}{c|}{CUB (\%)}
& \multicolumn{3}{c}{SCars (\%)} \\
\cmidrule(lr){2-4} \cmidrule(lr){5-7}
& All & Old & New & All & Old & New \\
\midrule
w/o StatReg
& 47.0 & 66.9 & 37.0 & 47.2 & 69.7 & 36.4 \\
w/o AngMargin
& 23.7 & 6.3 & 32.4 & 14.6 & 5.3 & 19.0 \\
w/o OldRefSel
& 48.0 & 63.9 & 40.1 & 47.3 & 72.2 & 35.3 \\
w/o EvidenceRouting
& 46.6 & 66.7 & 36.5 & 47.1 & 70.4 & 35.9 \\
w/o DynBirthTight
& 47.1 & 66.1 & 37.6 & 47.4 & 69.5 & 36.7 \\
\rowcolor{gray!10}
\textbf{\method~(full)}
& \textbf{49.7} & \textbf{67.1} & \textbf{41.0}
& \textbf{49.9} & \textbf{72.4} & \textbf{39.0} \\
\bottomrule
\end{tabular}
}
\vspace{-1.5em}
\end{table}
\vspace{-0.5em}
\subsection{Ablation study}
\textbf{The effectiveness of each module.}
Table~\ref{tab:ablation_cub_scars} demonstrates the individual contribution of each component within the \method{} framework under the Strict-Hungarian protocol. The most significant performance degradation occurs when the angular-margin objective is removed, causing "All" accuracy to plummet from 49.7\% to 23.7\% on CUB and from 49.9\% to 14.6\% on SCars. This confirms that establishing a well-defined spherical geometry is the fundamental prerequisite for effective category discovery. Furthermore, excluding statistical regularization consistently harms results across both datasets, verifying that proxy supervision on labeled data is vital for shaping a discriminative representation space suitable for later old-versus-new decisions. Regarding the inference-time decision modules, the removal of OldRefSel primarily degrades "Old" accuracy, while eliminating EvidenceRouting results in a notable decrease in "New" accuracy, validating its role in preventing premature competition between unstable novel prototypes and base classes. Finally, excluding DynBirthTight lowers the overall scores on both datasets, indicating that a dynamic birth boundary is crucial for suppressing redundant prototype instantiation. Together, these modules ensure that \method{} achieves the optimal trade-off between knowledge preservation and novel discovery without relying on a single dominant heuristic.

\begin{figure}[t]
  \centering
  \includegraphics[width=1\columnwidth]{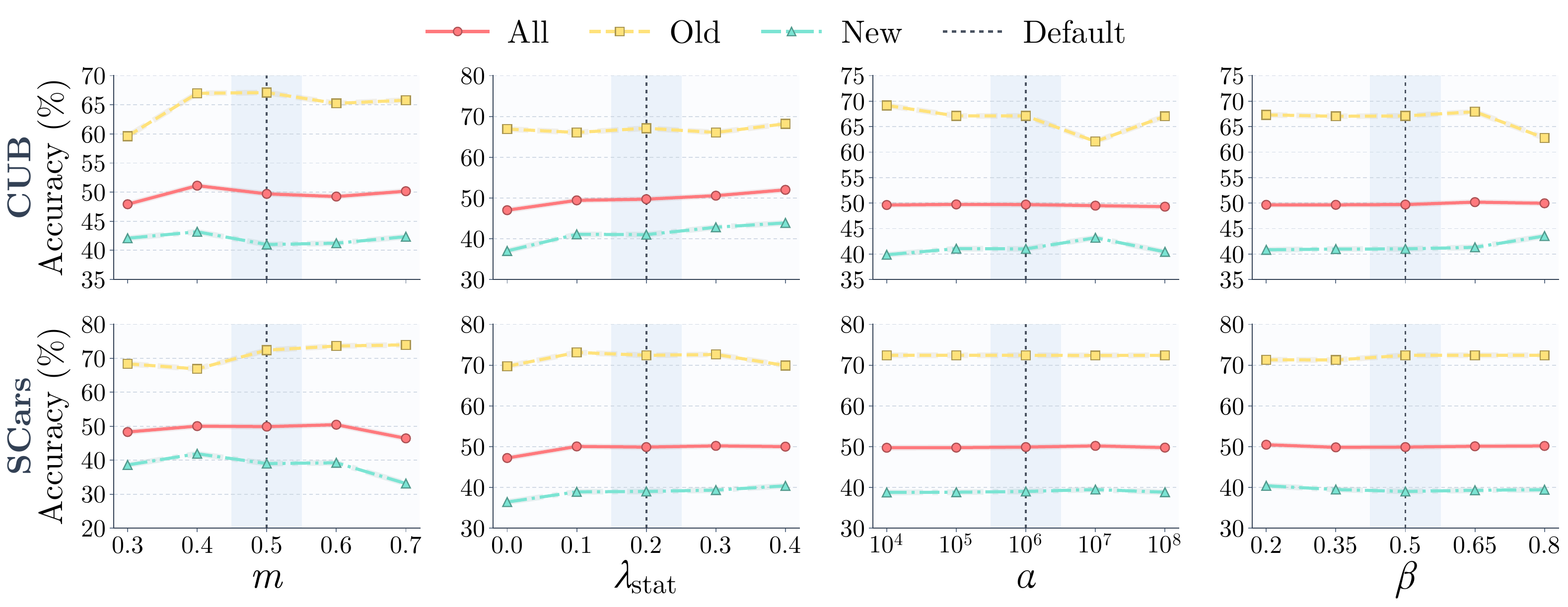}
  \vspace{-1em}
  \caption{
   Hyperparameter sensitivity analysis of $m, \lambda_{\text{stat}}, \alpha$, and $\beta$. \method{} exhibits robust performance across a wide range of settings, with default configurations (dashed lines) situated in stable regions for all metrics. While SCars New accuracy is sensitive to $m > 0.6$, the model remains remarkably stable across variations in $\lambda_{\text{stat}}, \alpha$, and $\beta$. These results validate the reliability and generalizability of our default parameter selections.
  }
  \Description{A two-row, four-column hyperparameter analysis figure. The top row is CUB and the bottom row is SCars. The four columns correspond to m, lambda stat, alpha, and beta. Each panel compares All, Old, New, and the default setting.}
  \label{fig:hparam_grid_non_p_nper}
  \vspace{-0.9em}
\end{figure}

\textbf{Hyperparameter Sensitivity Analysis.}
Fig.~\ref{fig:hparam_grid_non_p_nper} illustrates that \method{} maintains robust performance across a wide range of hyperparameter settings, with the default configurations lying in a stable region for all metrics. For the angular margin $m$, accuracy on CUB remains relatively flat after a mild peak near $m=0.5$, whereas SCars exhibits higher sensitivity, particularly in "New" accuracy, which drops significantly as $m$ exceeds 0.6. This suggests that while a clear margin is necessary to tighten classes, an overly aggressive value can distort the feature geometry for novel categories. In contrast, the model shows remarkable robustness to the statistical regularizer $\lambda_{\text{stat}}$, with "All" accuracy maintaining a nearly flat trajectory across various scales on both datasets. Similarly, performance stays stable across several orders of magnitude for the adaptation parameter $\alpha$ and a broad range for $\beta$, with only minor fluctuations in accuracy, further validating the reliability and generalizability of our default hyperparameter choices under diverse data distributions.

\begin{figure}[t]
  \centering
  \begin{minipage}[t]{0.5\columnwidth}
    \centering
    \includegraphics[width=\linewidth]{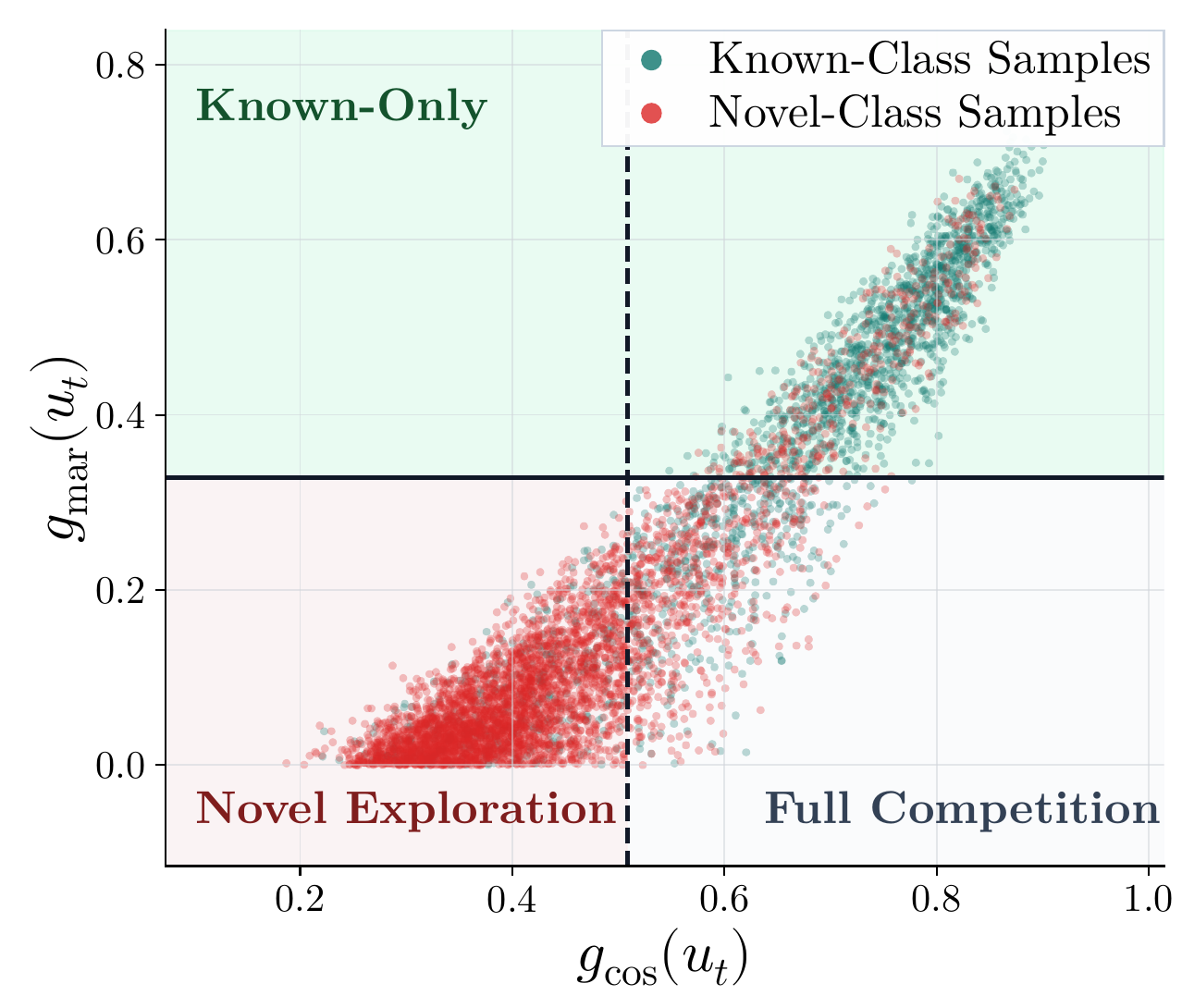}
    
    {\small (a) Routing Regions}
  \end{minipage}\hfill
  \begin{minipage}[t]{0.5\columnwidth}
    \centering
    \includegraphics[width=\linewidth]{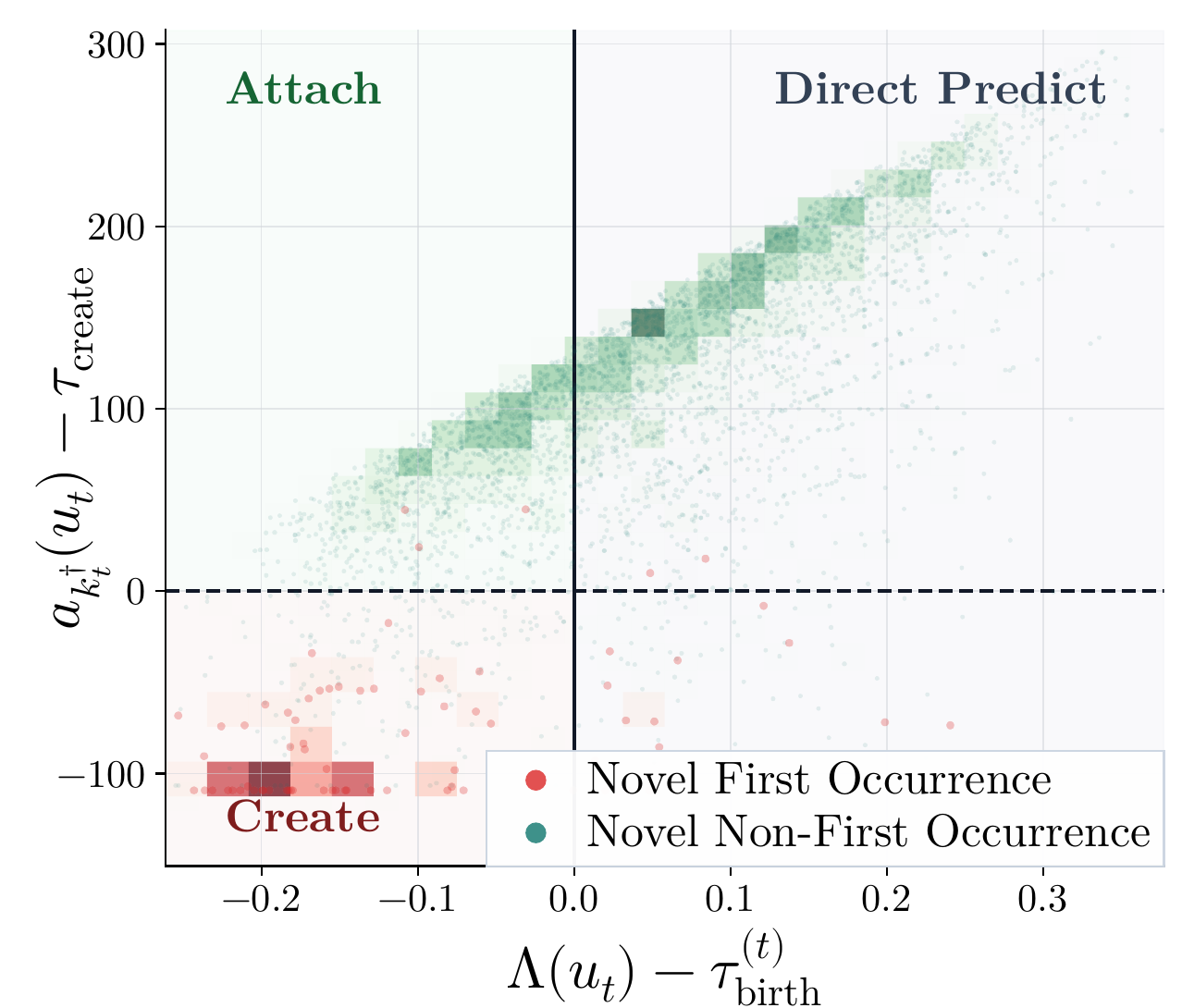}
    
    {\small (b) Attach/Create Regions}
  \end{minipage}
  \vspace{-0.2em}
\caption{Visualization of support-set calibration on Stanford Cars. (a) Routing Regions: The pair $(\tau_{\text{hi}}, \tau_{\text{lo}})$ partitions the stream into Known-Only, Novel Exploration, and Full Competition zones, reserving an uncertainty buffer for ambiguous samples. (b) Attach/Create Regions: Calibrated thresholds categorize samples into Direct Predict, Attach, and Create regions. The distribution shows that "Novel First Occurrences" (red dots) are correctly isolated for creation, while "Novel Non-First Occurrences" (green dots) are absorbed, effectively suppressing redundant prototype births.}
  \Description{A single-column figure with two side-by-side plots on Stanford Cars. The left plot shows routing regions induced by support-calibrated known-class evidence thresholds. The right plot shows direct-prediction, attach, and create regions induced by support-calibrated birth and creation thresholds.}
  \label{fig:scars_calibration_analysis}
    \vspace{-0.9em}
\end{figure}

\textbf{Visualization of Support-Set Calibration.}
A schematic geometric analysis provided in Fig.~\ref{fig:scars_calibration_analysis} clarifies how support-set calibration optimizes the inference process on the Stanford Cars dataset. As shown in Fig.~\ref{fig:scars_calibration_analysis}(a), the calibrated routing pair $(\tau_{\text{hi}}, \tau_{\text{lo}})$ successfully partitions the incoming stream into Known-Only, Novel Exploration, and Full Competition zones based on the strength of known-class evidence. This mechanism effectively reserves an uncertainty buffer for samples with weak evidence, reducing premature competition between unstable novel prototypes and strong base classes. Once samples enter the novel stage, the calibrated birth and creation thresholds accurately categorize them into Direct Prediction, Attach, or Create regions. As visualized in Fig.~\ref{fig:scars_calibration_analysis}(b), "Novel Non-First Occurrences" (green dots) correctly concentrate in the prediction and attach zones, while "Novel First Occurrences" (red dots) are pushed toward the creation region. This behavior confirms that \method{} suppresses unnecessary prototype births while effectively capturing genuinely new semantic concepts.

\begin{table}[tbp]
\small
\centering
\caption{Plug-and-play generalization of our support-set calibration (SSC) across existing OCD baselines. We evaluate the transferability of SSC by replacing the original threshold-fusion rule in AGE and the heuristic Hamming-radius rule in PHE-12bit  with our calibrated logic under the Strict-Hungarian protocol. $\Delta$ indicates the absolute performance gain over the corresponding original methods.}
\label{tab:transfer_ssc}
\resizebox{0.45\textwidth}{!}{
\begin{tabular}{l|ccc|ccc}
\toprule
\multirow{2}{*}[-0.6ex]{Method}
& \multicolumn{3}{c|}{CUB (\%)}
& \multicolumn{3}{c}{SCars (\%)} \\
\cmidrule(lr){2-4} \cmidrule(lr){5-7}
& All & Old & New & All & Old & New \\
\midrule
AGE~\cite{AGE}
& 46.3 & 59.8 & 39.4
& 34.8 & 62.7 & 21.3 \\
\midrule
\rowcolor{gray!10}
AGE+SSC
& 46.7 & 60.8 & 39.7
& 36.1 & 64.4 & 21.9 \\
\rowcolor{gray!10}
$\Delta$
& $\mathbf{\uparrow \best{0.4}}$ & $\mathbf{\uparrow \best{1.0}}$ & $\mathbf{\uparrow \best{0.3}}$
& $\mathbf{\uparrow \best{1.3}}$ & $\mathbf{\uparrow \best{1.7}}$ & $\mathbf{\uparrow \best{0.6}}$ \\
\midrule
PHE-12bit~\cite{PHE}
& 36.4 & 55.8 & 27.0
& 31.3 & 61.9 & 16.8 \\
\midrule
\rowcolor{gray!10}
PHE-12bit+SSC
& 39.1 & 60.6 & 28.3
& 34.2 & 66.7 & 17.9 \\
\rowcolor{gray!10}
$\Delta$
& $\mathbf{\uparrow \best{2.7}}$ & $\mathbf{\uparrow \best{4.8}}$ & $\mathbf{\uparrow \best{1.3}}$
& $\mathbf{\uparrow \best{2.9}}$ & $\mathbf{\uparrow \best{4.8}}$ & $\mathbf{\uparrow \best{1.1}}$ \\
\bottomrule
\end{tabular}
}
  \vspace{-1.0em}
\end{table}

\textbf{Plug-and-Play Generalization.}
To verify that our calibration logic is a model-agnostic inference-time layer, we transplant the Support-Set-Calibrated (SSC) principle into the representative OCD baselines AGE and PHE-12bit. As reported in Table~\ref{tab:transfer_ssc}, the integrated variants consistently outperform their original counterparts on both CUB and Stanford Cars. For AGE+SSC, we observe steady improvements across all metrics, including a $+1.3$ point gain in "All" accuracy and a $+1.7$ point gain in "Old" accuracy on SCars. The benefits are even more pronounced for PHE-12bit, where the addition of SSC logic yields a substantial $+4.8$ point increase in "Old" accuracy on both datasets, alongside a significant rise in "All" accuracy by $+2.7$ and $+2.9$ points, respectively. These results reinforce our central claim that support-set calibration is a reusable decision module that can be seamlessly incorporated into existing OCD pipelines to mitigate category explosion and enhance discovery stability regardless of the underlying representation.

\section{Conclusion}
We revisited on-the-fly category discovery from a decision-centric perspective and proposed \method, a support-set-calibrated framework for adaptive online discovery. Rather than relying on a fixed global threshold, \method{} models streaming inference as a hierarchy of decisions, including known-class routing, birth-aware novel assignment, and attach-versus-create operations over dynamic prototype memory. By aligning offline calibration with online inference through proxy tasks, and by continuously updating decision thresholds with mature novel prototypes, \method{} enables more stable and consistent category formation over time.

As a plug-and-play inference-time module, \method{} introduces no heavy training overhead and requires no dataset-specific tuning, yet yields significant gains over representative baselines across seven benchmarks. These findings highlight that the key to effective OCD lies not only in learning strong representations, but also in making well-calibrated and adaptive decisions throughout the discovery process.

\bibliographystyle{ACM-Reference-Format}
\bibliography{sample-base}

\clearpage
\appendix
\def\PACOCombinedAppendix{1}
\ifdefined\PACOCombinedAppendix
\newcommand{\mainsection}[1]{Sec.~\ref{#1}}
\newcommand{\mainequation}[1]{Eq.~\eqref{#1}}
\newcommand{\mainequations}[2]{Eqs.~\eqref{#1}--\eqref{#2}}
\newcommand{\maintable}[1]{Table~\ref{#1}}
\newcommand{\mainfigure}[1]{Fig.~\ref{#1}}
\newcommand{\mainalgorithm}[1]{Algorithm~\ref{#1}}
\else
\documentclass[sigconf, screen, nonacm, balance=false]{acmart}
\usepackage[table]{xcolor}
\usepackage{multirow}
\usepackage[ruled,vlined]{algorithm2e}
\usepackage{xr-hyper}

\AtBeginDocument{\providecommand\BibTeX{{Bib\TeX}}}

\newcommand{\citeplaceholder}{\textcolor{blue}{[cite]}}
\definecolor{ltgray}{gray}{0.93}
\newcommand{\best}[1]{\textcolor{red}{#1}}
\newcommand{\second}[1]{\textcolor{blue}{#1}}
\newcommand{\method}{PACO}
\SetKwInput{KwIn}{Input}
\SetKwInput{KwOut}{Output}
\SetKwInput{KwInit}{Init}

\settopmatter{printacmref=false}
\pagestyle{plain}

\externaldocument[main-]{sample-sigconf-authordraft}

\renewcommand{\theequation}{S\arabic{equation}}

\newcommand{\mainsection}[1]{Sec.~\ref{main-#1}}
\newcommand{\mainequation}[1]{Eq.~\eqref{main-#1}}
\newcommand{\mainequations}[2]{Eqs.~\eqref{main-#1}--\eqref{main-#2}}
\newcommand{\maintable}[1]{Table~\ref{main-#1}}
\newcommand{\mainfigure}[1]{Fig.~\ref{main-#1}}
\newcommand{\mainalgorithm}[1]{Algorithm~\ref{main-#1}}
\begin{document}

\title{Supplementary Material for PACO: Proxy-Task Alignment and Online Calibration for On-the-Fly Category Discovery}

\makeatletter
\gdef\authors{}
\gdef\shortauthors{}
\gdef\addresses{}
\makeatother

\maketitle
\setcounter{figure}{\numexpr\getrefnumber{main-fig:scars_calibration_analysis}\relax}
\setcounter{table}{\numexpr\getrefnumber{main-tab:transfer_ssc}\relax}
\fi

\section*{Appendix Overview}
This appendix is organized as follows. Sec.~\ref{sec:app_impl} summarizes the implementation details and default configuration. Sec.~\ref{sec:app_eval} specifies the evaluation protocols. Sec.~\ref{sec:app_additional_experiments} supplements the empirical results with transfer-study details and additional analyses. Sec.~\ref{sec:app_algorithm} gives an algorithmic summary of the complete offline-to-online pipeline. Sec.~\ref{sec:app_method_details} expands the method details omitted from the main paper. Sec.~\ref{sec:app_limitations} discusses the scope and limitations of \method{}.

\section{Implementation Details}
\label{sec:app_impl}
This section summarizes the dataset statistics and the default configuration used throughout the paper. We denote the labeled support set by $\mathcal{D}_S$, the online query stream by $\mathcal{D}_Q$, the full query label space by $\mathcal{Y}_Q$, the base-class label space by $\mathcal{Y}_S$, and the novel-class label space by $\mathcal{Y}_{novel}=\mathcal{Y}_Q\setminus\mathcal{Y}_S$.

\subsection{Dataset Statistics}
Following the standard OCD protocol described in \mainsection{subsec:exp_setup}, each dataset is first split into base and novel categories. For the base categories, $50\%$ of the training samples are used to construct the labeled support set $\mathcal{D}_S$. The remaining base-category samples together with all novel-category samples form the online query stream $\mathcal{D}_Q$.

\begin{table}[ht]
  \centering
  \small
  \setlength{\tabcolsep}{6pt}
  \renewcommand{\arraystretch}{1.08}
  \caption{Dataset statistics under the OCD protocol used in this paper.}
  \label{tab:appendix_dataset_stats}
  \begin{tabular}{lccccc}
    \toprule
    Dataset & $|\mathcal{Y}_Q|$ & $|\mathcal{Y}_S|$ & $|\mathcal{Y}_{novel}|$ & $|\mathcal{D}_S|$ & $|\mathcal{D}_Q|$ \\
    \midrule
    CIFAR-10 & 10 & 5 & 5 & 12,500 & 37,500 \\
    CIFAR-100 & 100 & 80 & 20 & 20,000 & 30,000 \\
    ImageNet-100 & 100 & 50 & 50 & 31,860 & 95,255 \\
    CUB-200-2011 & 200 & 100 & 100 & 1,498 & 4,496 \\
    Stanford Cars & 196 & 98 & 98 & 2,000 & 6,144 \\
    Oxford-IIIT Pet & 37 & 19 & 18 & 942 & 2,738 \\
    Food-101 & 101 & 51 & 50 & 19,125 & 56,625 \\
    \bottomrule
  \end{tabular}
\end{table}

\begin{table*}[t]
  \centering
  \small
  \setlength{\tabcolsep}{5pt}
  \renewcommand{\arraystretch}{1.04}
  \caption{Mean and standard deviation of \method{} over five independent random seeds under the Strict--Hungarian and Greedy--Hungarian protocols. Residual variation is concentrated mainly on New accuracy for the fine-grained benchmarks.}
  \label{tab:app_seed_errorbar}
  \begin{tabular}{l|ccc|ccc}
    \toprule
    \multirow{2}{*}{Dataset} & \multicolumn{3}{c|}{Strict--Hungarian (\%)} & \multicolumn{3}{c}{Greedy--Hungarian (\%)} \\
    \cmidrule(lr){2-4}\cmidrule(lr){5-7}
    & All & Old & New & All & Old & New \\
    \midrule
    CIFAR-10 & 58.07$\pm$0.57 & 57.60$\pm$0.86 & 58.30$\pm$0.94 & 87.13$\pm$0.20 & 97.40$\pm$0.09 & 82.00$\pm$0.29 \\
    CIFAR-100 & 67.03$\pm$0.48 & 78.80$\pm$0.67 & 43.50$\pm$1.24 & 71.10$\pm$0.35 & 83.60$\pm$0.50 & 46.10$\pm$0.98 \\
    ImageNet-100 & 67.81$\pm$0.34 & 92.70$\pm$0.29 & 55.30$\pm$0.81 & 68.87$\pm$0.27 & 92.70$\pm$0.23 & 56.90$\pm$0.69 \\
    CUB-200-2011 & 49.70$\pm$1.14 & 67.10$\pm$1.47 & 41.00$\pm$1.89 & 53.83$\pm$0.79 & 79.70$\pm$1.02 & 40.90$\pm$1.58 \\
    Stanford Cars & 49.87$\pm$1.06 & 72.40$\pm$1.28 & 39.00$\pm$2.05 & 53.07$\pm$0.71 & 79.10$\pm$0.91 & 40.50$\pm$1.43 \\
    Oxford-IIIT Pet & 71.86$\pm$1.88 & 67.40$\pm$2.76 & 74.20$\pm$3.08 & 78.15$\pm$1.11 & 85.70$\pm$1.46 & 74.20$\pm$1.98 \\
    Food-101 & 39.99$\pm$0.82 & 69.00$\pm$1.18 & 25.20$\pm$1.63 & 43.96$\pm$0.49 & 72.70$\pm$0.68 & 29.30$\pm$1.02 \\
    \midrule
    Average & 57.76$\pm$0.90 & 72.14$\pm$1.22 & 48.07$\pm$1.66 & 65.16$\pm$0.56 & 84.41$\pm$0.70 & 52.84$\pm$1.14 \\
    \bottomrule
  \end{tabular}
\end{table*}

\subsection{Default Configuration}
Unless otherwise specified, all experiments use a DINO-pretrained ViT-B/16 backbone together with a linear projection head of dimension $d=768$. The representation model is trained for 20 epochs with AdamW using a batch size of 128, a learning rate of $1\times 10^{-3}$ for the projection head, a smaller learning rate of $5\times 10^{-5}$ for the backbone, and weight decay $10^{-4}$. For the standardized spherical representation in \mainsection{subsec:representation}, the angular-margin classifier uses $s=30.0$ and $m=0.5$, and the support-only statistical regularization in Sec.~\ref{app:stat_reg} uses $\lambda_{\mathrm{stat}}=0.2$ and $\lambda_{\mathrm{cmp}}=1.0$. For the support-fixed geometry and online decision pipeline, we use the whitening constant $\epsilon=10^{-5}$, score temperature $T=1.0$, Dirichlet smoothing $\alpha=10^6$, micro-prototype exponent $\beta=0.5$, and support-spread coefficient $c_{\mathrm{spread}}=1.0$. These values are shared across all datasets so that the same support-set-calibrated decision pipeline can be applied without dataset-specific tuning. After support calibration, the base-class references remain fixed during streaming inference.

\textbf{Compute Environment.}
Unless otherwise specified, all experiments are conducted on a server equipped with Intel Xeon Platinum 8352Y CPUs and a single NVIDIA A800-SXM4-80GB GPU.

\section{Evaluation Protocol Details}
\label{sec:app_eval}
Following \mainsection{subsec:exp_setup}, once the stream has been processed, we keep the largest $|\mathcal{Y}_Q|$ predicted clusters by size. Let $\widehat{\mathcal{C}}$ denote the retained predicted cluster indices, and keep $\hat y_j\in\widehat{\mathcal{C}}$ as the retained predicted cluster index of sample $j$. The Hungarian step relabels retained predicted cluster indices to ground-truth labels before accuracy is computed. In this section only, we use Old and New to match the evaluation terminology in the main paper. Let
\begin{equation}
\mathcal{I}_{\mathrm{old}}=\{j:y_j\in\mathcal{Y}_S\},\qquad
\mathcal{I}_{\mathrm{new}}=\{j:y_j\in\mathcal{Y}_{novel}\}.
\label{eq:app_eval_index_sets}
\end{equation}
Let $\Pi(A,B)$ denote the set of one-to-one assignments from $A$ to $B$. Unless otherwise specified, Strict--Hungarian is our primary evaluation metric, because it uses a single global matching over the full query stream and therefore provides the more stringent assessment of online discovery quality.

\textbf{Strict-Hungarian.}
Under the Strict-Hungarian protocol, a single global assignment is computed over the full query stream:
\begin{equation}
\pi^\star
=
\arg\max_{\pi\in\Pi(\widehat{\mathcal{C}},\mathcal{Y}_Q)}
\sum_{j=1}^{N_Q}\mathbb{I}[\pi(\hat y_j)=y_j].
\label{eq:app_eval_strict_match}
\end{equation}
The reported accuracies are
\begin{equation}
\mathrm{ACC}^{\mathrm{S}}_{\mathrm{All}}
=
\frac{1}{N_Q}
\sum_{j=1}^{N_Q}\mathbb{I}[\pi^\star(\hat y_j)=y_j],
\label{eq:app_eval_strict_all}
\end{equation}
\begin{equation}
\mathrm{ACC}^{\mathrm{S}}_{\mathrm{Old}}
=
\frac{1}{|\mathcal{I}_{\mathrm{old}}|}
\sum_{j\in\mathcal{I}_{\mathrm{old}}}\mathbb{I}[\pi^\star(\hat y_j)=y_j].
\label{eq:app_eval_strict_old}
\end{equation}
\begin{equation}
\mathrm{ACC}^{\mathrm{S}}_{\mathrm{New}}
=
\frac{1}{|\mathcal{I}_{\mathrm{new}}|}
\sum_{j\in\mathcal{I}_{\mathrm{new}}}\mathbb{I}[\pi^\star(\hat y_j)=y_j].
\label{eq:app_eval_strict_new}
\end{equation}
This protocol uses one shared assignment for the entire stream and therefore penalizes both old-class fragmentation and cross-contamination between old and novel predictions.

\textbf{Greedy-Hungarian.}
Under the Greedy-Hungarian protocol, the old and novel subsets are matched independently. Let
\begin{equation}
\widehat{\mathcal{C}}_{\mathrm{old}}=\{\hat y_j:j\in\mathcal{I}_{\mathrm{old}}\},\qquad
\widehat{\mathcal{C}}_{\mathrm{new}}=\{\hat y_j:j\in\mathcal{I}_{\mathrm{new}}\}.
\end{equation}
\noindent The corresponding assignments are
\begin{equation}
\pi^\star_{\mathrm{old}}
=
\arg\max_{\pi\in\Pi(\widehat{\mathcal{C}}_{\mathrm{old}},\mathcal{Y}_S)}
\sum_{j\in\mathcal{I}_{\mathrm{old}}}\mathbb{I}[\pi(\hat y_j)=y_j],
\label{eq:app_eval_greedy_match_old}
\end{equation}
\begin{equation}
\pi^\star_{\mathrm{new}}
=
\arg\max_{\pi\in\Pi(\widehat{\mathcal{C}}_{\mathrm{new}},\mathcal{Y}_{novel})}
\sum_{j\in\mathcal{I}_{\mathrm{new}}}\mathbb{I}[\pi(\hat y_j)=y_j].
\label{eq:app_eval_greedy_match_new}
\end{equation}
The corresponding subset accuracies are
\begin{equation}
\mathrm{ACC}^{\mathrm{G}}_{\mathrm{Old}}
=
\frac{1}{|\mathcal{I}_{\mathrm{old}}|}
\sum_{j\in\mathcal{I}_{\mathrm{old}}}\mathbb{I}[\pi^\star_{\mathrm{old}}(\hat y_j)=y_j].
\label{eq:app_eval_greedy_old}
\end{equation}
\begin{equation}
\mathrm{ACC}^{\mathrm{G}}_{\mathrm{New}}
=
\frac{1}{|\mathcal{I}_{\mathrm{new}}|}
\sum_{j\in\mathcal{I}_{\mathrm{new}}}\mathbb{I}[\pi^\star_{\mathrm{new}}(\hat y_j)=y_j].
\label{eq:app_eval_greedy_new}
\end{equation}
and the reported All accuracy is their size-weighted combination:
\begin{equation}
\mathrm{ACC}^{\mathrm{G}}_{\mathrm{All}}
=
\frac{|\mathcal{I}_{\mathrm{old}}|}{N_Q}\mathrm{ACC}^{\mathrm{G}}_{\mathrm{Old}}
+
\frac{|\mathcal{I}_{\mathrm{new}}|}{N_Q}\mathrm{ACC}^{\mathrm{G}}_{\mathrm{New}}.
\label{eq:app_eval_greedy_all}
\end{equation}
This makes explicit that Greedy-Hungarian evaluates the old and novel subsets separately, whereas Strict-Hungarian uses a single global matching over the full query stream.

\section{Additional Experiments and Analysis}
\label{sec:app_additional_experiments}

\subsection{Plug-and-Play Transfer Details}
The plug-and-play transfer study in \maintable{tab:transfer_ssc} changes only the inference-time decision module of the host method. The goal is to isolate the effect of the support-set-calibrated (SSC) decision logic while leaving the representation and memory components of the original method intact.

\textbf{AGE+SSC.}
For AGE, we keep its original feature representation, category-density modeling, and online memory update unchanged. The only modification is to replace the original threshold-fusion step with the support-set-calibrated decision logic of \method{}. In this way, AGE+SSC preserves the host method's representation and density evolution while recalibrating the known-versus-novel transition with support-set statistics.

\textbf{PHE-12bit+SSC.}
For PHE-12bit, we keep the original 12-bit hash encoder, hash prototype construction, and online prototype update unchanged. We replace only the heuristic Hamming-radius attach-versus-create rule with the support-set-calibrated boundary estimated from pseudo-novel support replays. No additional backbone retraining, representation replacement, or dataset-specific threshold search is introduced in this transfer.

\subsection{Error Bars Across Random Seeds}
To assess the sensitivity of \method{} to random seeds, Table~\ref{tab:app_seed_errorbar} reports the mean and standard deviation over five independent runs under both the Strict--Hungarian and Greedy--Hungarian protocols. Across datasets, the same overall performance pattern as in the main paper is preserved, and the average standard deviation remains limited at $0.90/1.22/1.66$ for All/Old/New under Strict--Hungarian and $0.56/0.70/1.14$ under Greedy--Hungarian. The remaining variation is more visible on New accuracy, especially on fine-grained benchmarks, which is expected in a single-pass and path-dependent OCD stream. Taken together, these results indicate that the performance of \method{} remains reasonably stable across random seeds.

\subsection{Sensitivity to the Support Ratio}
\begin{figure}[t]
  \centering
  \begin{minipage}[t]{0.333\columnwidth}
    \centering
    \includegraphics[width=\linewidth]{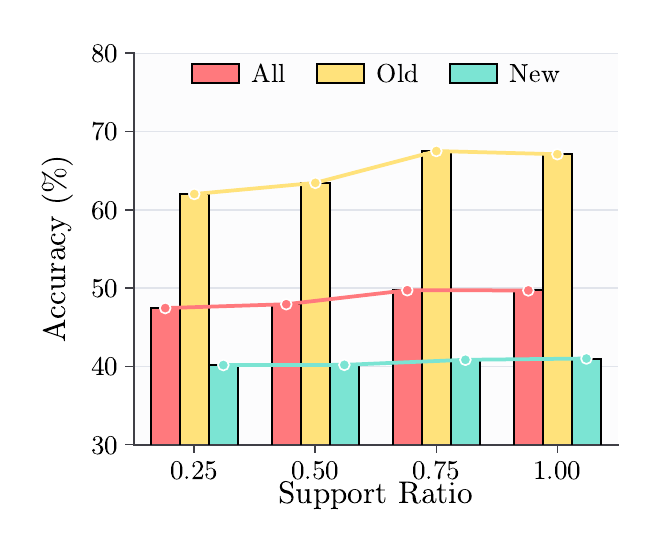}

    {\small (a) CUB-200-2011}
  \end{minipage}\hfill
  \begin{minipage}[t]{0.333\columnwidth}
    \centering
    \includegraphics[width=\linewidth]{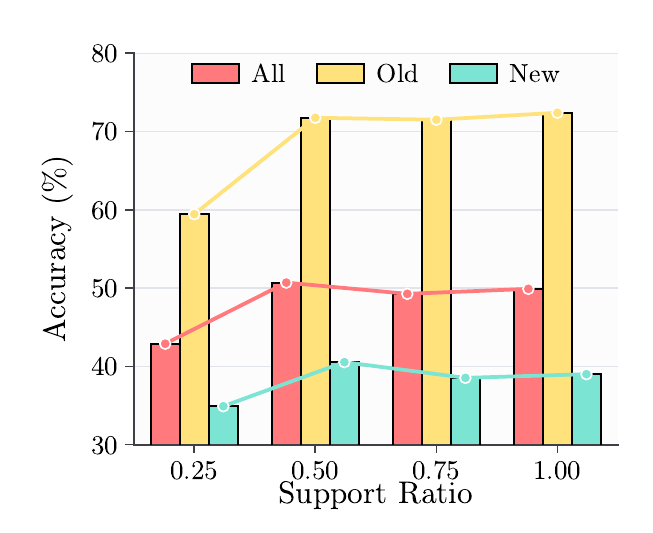}

    {\small (b) Stanford Cars}
  \end{minipage}\hfill
  \begin{minipage}[t]{0.333\columnwidth}
    \centering
    \includegraphics[width=\linewidth]{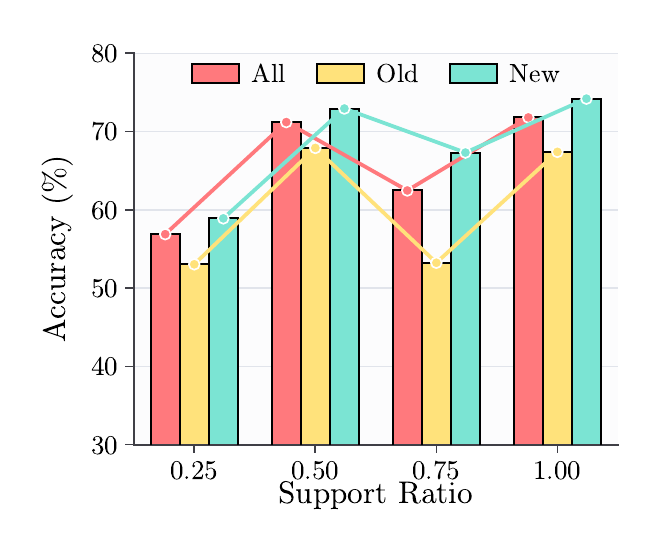}

    {\small (c) Oxford-IIIT Pet}
  \end{minipage}
  \caption{Strict--Hungarian All/Old/New accuracy under different support ratios}
  \Description{Three side-by-side plots showing Strict-Hungarian All, Old, and New accuracy against the support ratio for CUB-200-2011, Stanford Cars, and Oxford-IIIT Pet.}
  \label{fig:app_support_ratio_accuracy}
  \vspace{-1em}
\end{figure}

We next examine how the amount of support data available to the calibration stage affects Strict--Hungarian accuracy. In Fig.~\ref{fig:app_support_ratio_accuracy}, the support ratio denotes the retained fraction of the default support set $\mathcal{D}_S$ and varies from $0.25$ to $1.00$ across three fine-grained benchmarks. This analysis is tied directly to the role of $\mathcal{D}_S$ in the support-set statistics of \mainequation{eq:support_stats} and \mainequation{eq:std_embed}, and in the proxy-task calibration of $\tau_{\text{hi}}$, $\tau_{\text{lo}}$, $\tau_{\text{birth}}$, and $\tau_{\text{create}}$ in \mainsection{subsec:calibration}. Across the three panels, larger support ratios generally improve Strict--Hungarian All and Old accuracy, while the response of Strict--Hungarian New accuracy remains more moderate and does not show a systematic degradation. Some local fluctuations remain, which is expected in single-pass OCD because early decisions can influence later outcomes. Overall, however, the trend suggests that larger support budgets improve the reliability of support-set calibration rather than preserving old classes at the expense of new-class discovery.

\subsection{Dynamic Birth Tightening over the Stream}
\begin{figure}[t]
  \centering
  \begin{minipage}[t]{0.49\columnwidth}
    \centering
    \includegraphics[width=\linewidth]{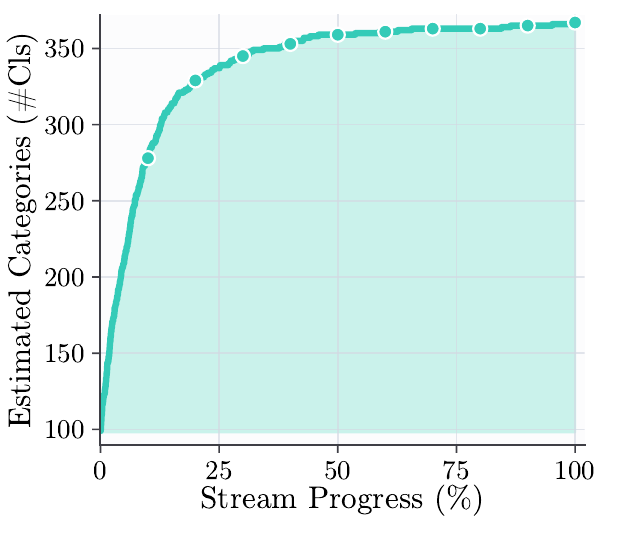}

    {\small (a) Estimated Categories}
  \end{minipage}\hfill
  \begin{minipage}[t]{0.49\columnwidth}
    \centering
    \includegraphics[width=\linewidth]{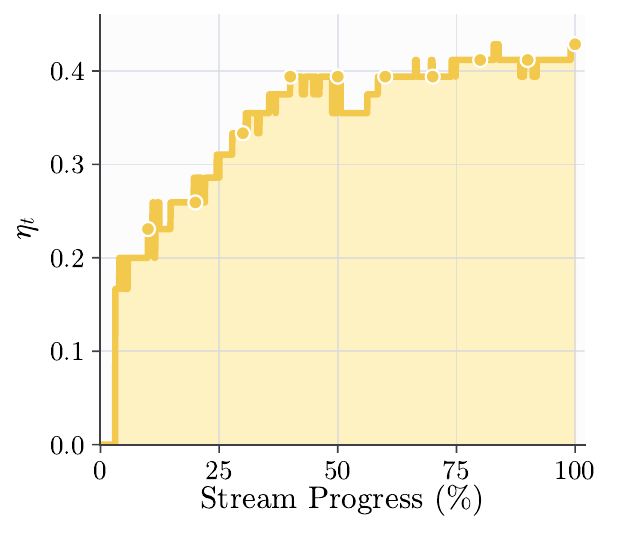}

    {\small (b) Mixing Coefficient}
  \end{minipage}
    \caption{Stream-level traces of the adaptive birth update on Stanford Cars under the default setting. (a) The estimated category count rises rapidly early in the stream and then flattens, indicating that most prototype creation occurs before the novel bank stabilizes. (b) The mixing coefficient $\eta_t$ increases as reliable novel structure accumulates, so the adaptive term in Eqs.~\eqref{eq:app_birth_dynamic} becomes more influential later in the stream. Markers denote $10\%$ stream increments.}
  \Description{Two side-by-side process plots on Stanford Cars. The first panel shows estimated category count over stream progress. The second panel shows the mixing coefficient eta over stream progress, with markers placed at ten-percent increments.}
  \label{fig:app_scars_birth_progress}
\end{figure}

\mainfigure{fig:scars_calibration_analysis} provides a static view of the calibrated decision regions. Fig.~\ref{fig:app_scars_birth_progress} shows how the adaptive birth update evolves over the Stanford Cars stream under the default setting. As shown in Fig.~\ref{fig:app_scars_birth_progress}(a), prototype creation is concentrated in the early stage of the stream: by $10\%$ progress, the memory has already expanded to 278 categories, whereas the remaining $90\%$ of the stream adds only 89 more. The trajectory then flattens substantially, indicating a transition from rapid novel expansion to a more stable regime. Over the same interval, Fig.~\ref{fig:app_scars_birth_progress}(b) shows that $\eta_t$ increases gradually, so the adaptive term in Eqs.~\eqref{eq:app_birth_dynamic} receives more weight only after reliable novel structure has formed. Together with the flattening category trajectory in Fig.~\ref{fig:app_scars_birth_progress}(a), this pattern is consistent with the adaptive birth rule mainly affecting the later stage of the stream rather than the initial expansion required to instantiate novel categories.

\subsection{Wall-Clock Cost of Support-Set Calibration}
We further quantify the one-time wall-clock overhead introduced by the support-set-calibrated decision layer of \method{}. Specifically, we measure the offline support-set calibration stage that is executed after training and before the online query stream begins. Following the decomposition implied by Algorithm~\ref{alg:paco_appendix}, we report the cost of support feature extraction on $\mathcal{D}_S$, the cost of support-only calibration of the routing, birth, and create thresholds, and their sum. All numbers are measured under the default implementation of \method{} and are reported as the mean and standard deviation over five runs.

\begin{table}[t]
  \centering
  \small
  \setlength{\tabcolsep}{3.5pt}
  \renewcommand{\arraystretch}{1.05}
  \caption{Wall-clock time of the one-time support-set calibration stage that initializes \method{} before online inference. For each dataset, we report the mean and standard deviation over five runs. The total cost is decomposed into support feature extraction and support-only calibration, and excludes both offline training and online stream processing.}
  \label{tab:app_support_calibration_time}
  \resizebox{\columnwidth}{!}{
  \begin{tabular}{lccc}
    \toprule
    Dataset & \shortstack{Total\\(s)} & \shortstack{Support feature\\extraction (s)} & \shortstack{Support-only\\calibration (s)} \\
    \midrule
    CIFAR-10 & 55.06$\pm$4.10 & 27.84$\pm$0.07 & 27.23$\pm$4.07 \\
    CIFAR-100 & 118.07$\pm$19.88 & 44.03$\pm$0.02 & 74.04$\pm$19.87 \\
    ImageNet-100 & 186.00$\pm$18.19 & 70.42$\pm$0.32 & 115.58$\pm$18.15 \\
    CUB-200-2011 & 8.48$\pm$0.58 & 4.17$\pm$0.07 & 4.31$\pm$0.63 \\
    Stanford Cars & 13.06$\pm$3.05 & 7.39$\pm$1.12 & 5.67$\pm$2.06 \\
    Oxford-IIIT Pet & 4.31$\pm$0.07 & 2.93$\pm$0.04 & 1.38$\pm$0.06 \\
    Food-101 & 110.54$\pm$21.97 & 42.58$\pm$0.18 & 67.96$\pm$22.01 \\
    \bottomrule
  \end{tabular}
  }
\end{table}

As shown in Table~\ref{tab:app_support_calibration_time}, the additional overhead remains a one-time pre-stream cost and scales mainly with the size of the support set. Small and medium-sized benchmarks incur only a few seconds, whereas CIFAR-100, Food-101, and ImageNet-100 require longer calibration because both support feature extraction and the support-set proxy computations operate over substantially larger support sets. In these measurements, support feature extraction is executed on a single NVIDIA A800-SXM4-80GB GPU, whereas the support-only calibration stage is executed on the host Intel Xeon Platinum 8352Y CPU. Importantly, this overhead is incurred once before the stream starts and therefore does not alter the single-pass online evaluation protocol used throughout the paper.

\begin{figure}[t]
  \centering
  \includegraphics[width=\columnwidth]{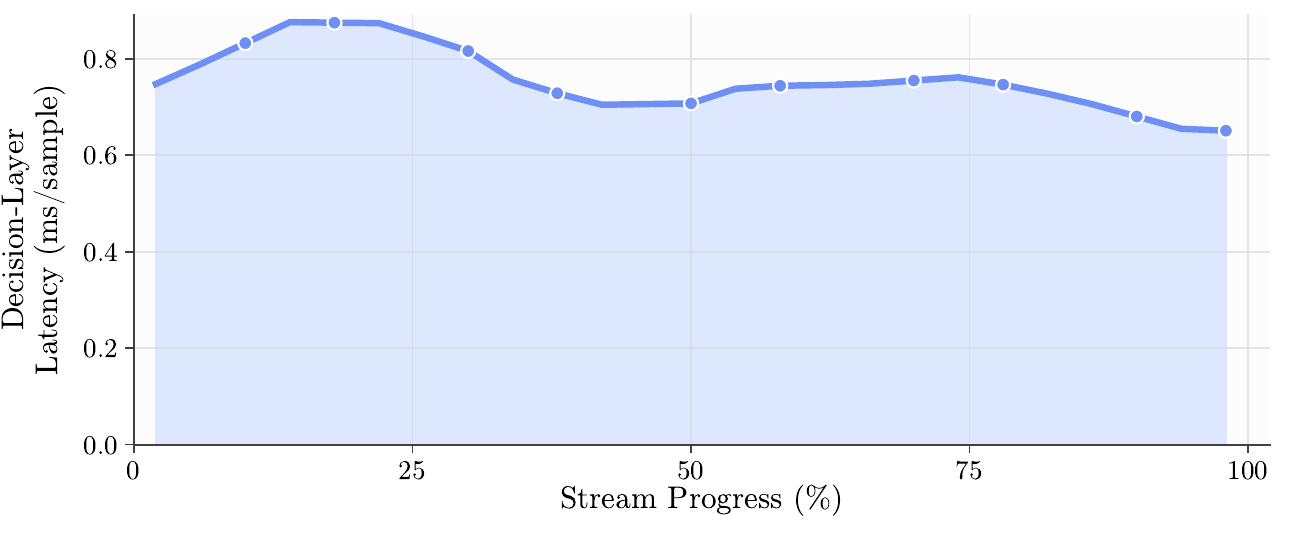}
  \caption{Per-sample latency of the inference-time decision layer of \method{} on Stanford Cars. Measurements use cached projected stream features, excluding backbone and projection-head extraction while retaining whitening and the downstream decision-stage operations. The curve is estimated from $4\%$ stream-progress bins, with markers placed at $10\%$ progress intervals. The latency remains below $1$ ms for most of the stream and does not exhibit sustained growth as the active prototype bank evolves.}
  \Description{A wide line plot showing decision-layer latency per sample against stream progress on Stanford Cars. The curve remains in a narrow band for most of the stream, with no sustained growth over time.}
  \label{fig:app_scars_decision_latency}
\end{figure}

\begin{algorithm}[t]
\scriptsize
\DontPrintSemicolon
\caption{\method: offline calibration and online inference}
\label{alg:paco_appendix}
\KwIn{Labeled support set $\mathcal{D}_S$; online query stream $\mathcal{D}_Q$; backbone $f_\theta$; projection head $g_\phi$; hyperparameters (e.g., $s$, $m$, $\lambda_{\mathrm{stat}}$, et al.).}
\KwOut{Streaming predictions $\{\hat y_t\}_{t=1}^{N_Q}$ and the final online memory state.}
\tcp{Offline support stage}
Train $(f_\theta,g_\phi)$ by minimizing Eqs.~\eqref{eq:app_total_train_loss}\;
Compute $(\mathbf{m}_S,\boldsymbol{\nu}_S)$ via \mainequation{eq:support_stats} and standardize support embeddings with \mainequation{eq:std_embed}\;
Build candidate base references by Eqs.~\eqref{eq:app_old_proto}--\eqref{eq:app_whitened_classifier} and retain $\{\boldsymbol{\psi}_k\}_{k=1}^{K_{\mathrm{base}}}$ by the support-set selection rule in Sec.~\ref{app:old_refsel}\;
Estimate $(\tau_{\mathrm{hi}},\tau_{\mathrm{lo}})$ from the routing proxy task and estimate $(\tau_{\mathrm{birth}},\tau_{\mathrm{create}})$ from the birth/create proxy tasks in \mainsection{subsec:calibration}\;
Set $\tau_{\mathrm{birth}}^{\mathrm{sup}}$ by Eqs.~\eqref{eq:app_birth_sup_shrink}; set $\tau_{\mathrm{birth}}^{(1)}\leftarrow\tau_{\mathrm{birth}}^{\mathrm{sup}}$; initialize $K_t\leftarrow K_{\mathrm{base}}$ and $\{n_k(1)\}_{k=1}^{K_{\mathrm{base}}}$ from support-class sizes\;
\tcp{Online stream}
\ForEach{incoming sample $x_t$ in $\mathcal{D}_Q$}{
Extract $h_t=g_\phi(f_\theta(x_t))$ and compute the standardized embedding $u_t$ with \mainequation{eq:std_embed}\;
Set $\mathcal{N}_t\leftarrow\{K_{\mathrm{base}}+1,\dots,K_t\}$ and compute the memory score $s_k(u_t)$ by \mainequation{eq:memory_score}, using $\boldsymbol{\psi}_k$ for $k\le K_{\mathrm{base}}$ and $\mu_k(t)$ otherwise\;
Compute the base-side evidence and route $C_t$ by \mainequation{eq:candidate_route}\;
\eIf{$C_t=\{1,\dots,K_{\mathrm{base}}\}$}{
$\hat y_t\leftarrow\arg\max_{k\in C_t}s_k(u_t)$\;
}{
\eIf{$C_t=\emptyset$}{
$\hat y_t\leftarrow K_t+1$\;
}{
Evaluate the birth statistic $\Lambda(u_t)$ by \mainequation{eq:birth_stat}\;
\eIf{$\Lambda(u_t)\ge\tau_{\mathrm{birth}}^{(t)}$}{
$\hat y_t\leftarrow\arg\max_{k\in C_t}s_k(u_t)$\;
}{
\eIf{$\mathcal{N}_t=\emptyset$}{
$\hat y_t\leftarrow K_t+1$\;
}{
Compute $\bar r_k(t)$, $\kappa_k(t)$, and $a_k(u_t)$ via Eqs.~\eqref{eq:app_attach_mean_resultant}--\eqref{eq:app_attach_score}\;
Obtain $k_t^\dagger$ from Eqs.~\eqref{eq:app_attach_argmax} and decide $\hat y_t$ by \mainequation{eq:attach_decision}\;
}
}
}
}
\eIf{$\hat y_t\le K_{\mathrm{base}}$}{
$n_{\hat y_t}(t+1)\leftarrow n_{\hat y_t}(t)+1$ and keep the base reference fixed\;
}{
\eIf{$\hat y_t=K_t+1$}{
Initialize the new prototype with Eqs.~\eqref{eq:app_attach_new_init} and set $K_t\leftarrow K_t+1$\;
}{
Update the selected novel prototype with Eqs.~\eqref{eq:app_attach_existing_update} and \eqref{eq:app_attach_existing_mu_update}\;
}
}
Build the updated mature novel set $\mathcal{M}_{t+1}$ by the rule in Eqs.~\eqref{eq:app_birth_mature_set}\;
\eIf{$|\mathcal{M}_{t+1}|<2$}{
$\tau_{\mathrm{birth}}^{(t+1)}\leftarrow\tau_{\mathrm{birth}}^{\mathrm{sup}}$\;
}{
Set $m_{t+1}\leftarrow\mathrm{median}(\{n_k(t+1):k\in\mathcal{M}_{t+1}\})$\;
Compute $\tau_{\mathrm{birth}}^{\mathrm{bank}}(t+1)$ and $\eta_{t+1}$ from the updated mature set by Eqs.~\eqref{eq:app_birth_bank}--\eqref{eq:app_birth_eta}, then set $\tau_{\mathrm{birth}}^{(t+1)}$ by the same rule as Eqs.~\eqref{eq:app_birth_dynamic}\;
}
}
\end{algorithm}

\subsection{Online Decision-Layer Latency During Streaming Inference}
Whereas Table~\ref{tab:app_support_calibration_time} quantifies the one-time pre-stream cost of support-set calibration, this subsection examines the per-sample latency of the inference-time decision layer during streaming inference. The active prototype bank evolves throughout the stream, which makes the incremental post-feature overhead the relevant quantity after support-set calibration has been completed. To isolate this cost from representation extraction, we measure the stage after cached projected stream features are available, while retaining whitening to standardized embeddings together with candidate routing, birth evaluation, attach-versus-create scoring, prototype-memory update, and dynamic birth-threshold update. Fig.~\ref{fig:app_scars_decision_latency} reports the resulting profile on Stanford Cars under the default configuration. The per-sample cost remains below $1$ ms for most of the stream and does not exhibit sustained growth as the active prototype bank evolves. Some local variation remains visible, which is consistent with the dependence of the executed decision path on the current candidate set and the state of the novel bank. The overall range nevertheless stays narrow over most of the stream, indicating that \method{} adds bounded incremental online overhead after projected features have been computed.

\section{Offline-to-Online Algorithm}
\label{sec:app_algorithm}
The complete offline-to-online pipeline of \method{} is summarized in Algorithm~\ref{alg:paco_appendix}.

\section{Additional Method Details}
\label{sec:app_method_details}
This section expands components that are already part of the main method but are omitted there for space: the support-set proxy regularization behind the training--inference alignment claim, the support-side geometry used by the routing and memory scores, and the dynamic birth-threshold refinement used during streaming inference. The subsection order follows the same offline-to-online narrative as \mainsection{sec:method}. To make the derivations easier to read, this supplementary document uses boldface for vector quantities while keeping the same symbols as in the main paper unless a local auxiliary symbol is introduced explicitly. The goal is not to introduce new stages after the fact, but to make the support-set-calibrated decision framework auditable end to end.

\subsection{Support-Only Statistical Episode Regularization}
\label{app:stat_reg}
This subsection spells out the proxy-task component behind the statistical regularization term mentioned in \mainsection{subsec:representation}. In addition to the additive angular-margin classification loss, training uses a support-only statistical regularization term that exposes the model to a proxy seen-versus-pseudo-unseen separation task. For each batch, we randomly split the appearing classes into a visible subset $\mathcal{V}$ and a hidden subset $\mathcal{H}$, with at least two visible classes and at least one hidden class. One augmented view is used to build visible-class prototypes, while the other view is used for evaluation. In this subsection only, $\mathbf{u}_i^{\mathrm{sup}}$ denotes the unit-norm projected feature of sample $i$ from the support-side augmented view; it is a local training-time symbol and is distinct from the inference-time standardized embedding $u$ in \mainequation{eq:std_embed}.

For each visible class $k$, the prototype is

\begin{equation}
\mathbf{p}_k
=
\frac{\sum_{i:y_i=k}\mathbf{u}_i^{\mathrm{sup}}}
{\left\|\sum_{i:y_i=k}\mathbf{u}_i^{\mathrm{sup}}\right\|_2}.
\label{eq:app_stat_proto}
\end{equation}

Visible query samples are evaluated against this visible prototype bank using paired leave-one-out construction. Hidden samples are evaluated against the same visible bank and therefore play the role of pseudo-unseen examples. This yields two families of statistics: the maximum prototype response and the top-1 to top-2 prototype margin.

We compare visible and hidden responses with the ranking operator

\begin{equation}
\mathcal{R}(\mathcal{A},\mathcal{B})
=
\frac{1}{|\mathcal{A}||\mathcal{B}|}
\sum_{a\in\mathcal{A}}
\sum_{b\in\mathcal{B}}
\mathrm{softplus}(b-a).
\label{eq:app_stat_rank}
\end{equation}

The response-ranking loss and margin-ranking loss are

\begin{equation}
\begin{aligned}
\mathcal{L}_{\mathrm{cos}}
=
\mathcal{R}(\{r_i^{+}\},\{r_j^{-}\}),
\qquad
\mathcal{L}_{\mathrm{margin}}
=
\mathcal{R}(\{m_i^{+}\},\{m_j^{-}\}),
\end{aligned}
\label{eq:app_stat_rank_losses}
\end{equation}

where $r_i^{+}$ and $m_i^{+}$ denote the maximum response and margin of visible samples, and $r_j^{-}$ and $m_j^{-}$ denote the corresponding statistics of hidden samples. We further add a compactness term

\begin{equation}
\mathcal{L}_{\mathrm{compact}}
=
\frac{1}{N_{\mathrm{visible}}}
\sum_i(1-\cos_{\mathrm{true},i}),
\label{eq:app_stat_compact}
\end{equation}

and form the statistical regularization

\begin{equation}
\mathcal{L}_{\mathrm{stat}}
=
\mathcal{L}_{\mathrm{cos}}
+
\mathcal{L}_{\mathrm{margin}}
+
\lambda_{\mathrm{cmp}}\mathcal{L}_{\mathrm{compact}}.
\label{eq:app_stat_reg_loss}
\end{equation}

The total training loss is

\begin{equation}
\mathcal{L}
=
\mathcal{L}_{\mathrm{cls}}
+
\lambda_{\mathrm{stat}}\mathcal{L}_{\mathrm{stat}}.
\label{eq:app_total_train_loss}
\end{equation}

This term is still trained only on support data, but it explicitly encourages the embedding space to separate confident base-class responses from pseudo-unseen ones before online inference begins.

\subsection{Support-Fixed Geometry and Base-Class Reference Selection}
\label{app:old_refsel}
This subsection expands the support-side geometry used by the routing and memory scores in \mainsection{subsec:inference}. After training, we re-extract support features and build support-fixed geometry. Using the support mean $\mathbf{m}_S$ and variance $\boldsymbol{\nu}_S$, we whiten and normalize support features in the same way as in the main paper. For each base class $k$, we compute its support prototype from the whitened unit features:

\begin{equation}
\boldsymbol{\mu}_k^{\mathrm{base}}
=
\frac{\sum_{i:y_i=k}\mathbf{u}_i}
{\left\|\sum_{i:y_i=k}\mathbf{u}_i\right\|_2}.
\label{eq:app_old_proto}
\end{equation}

At the same time, we also whiten each classifier direction:

\begin{equation}
\bar{\mathbf{w}}_k
=
\frac{\mathbf{w}_k\odot(\boldsymbol{\nu}_S+\epsilon)^{-1/2}}
{\|\mathbf{w}_k\odot(\boldsymbol{\nu}_S+\epsilon)^{-1/2}\|_2}.
\label{eq:app_whitened_classifier}
\end{equation}

The actual base-class reference direction $\boldsymbol{\psi}_k$ is selected from these two candidates by support-only criteria. We first compare support top-1 accuracy. If the two choices tie, we further compare their average top-1 to top-2 margin. The direction that performs better under this support-only check is retained as $\boldsymbol{\psi}_k$. This procedure lets the base-class side use whichever geometry is more reliable for the current support set instead of committing in advance to either support prototypes or classifier weights.

\subsection{Uniform Background Density and the vMF Interpretation}
\label{app:vmf}
This subsection justifies the birth statistic already used in \mainequation{eq:birth_stat}; it does not introduce a new probabilistic component beyond the main pipeline. The birth statistic in \mainequation{eq:birth_stat} compares the best available prototype explanation against a uniform background on the unit sphere. If the embedding dimension is $d$, then the standardized embeddings lie on $S^{d-1}$, whose surface area is

\begin{equation}
\mathrm{Area}(S^{d-1})=\frac{2\pi^{d/2}}{\Gamma(d/2)}.
\label{eq:app_vmf_area}
\end{equation}

Therefore, the uniform background density with respect to surface measure is

\begin{equation}
p_0=\frac{1}{\mathrm{Area}(S^{d-1})}
=
\frac{\Gamma(d/2)}{2\pi^{d/2}}.
\label{eq:app_vmf_p0}
\end{equation}

This quantity depends only on the embedding dimension and is computed once before streaming inference starts. In practice we use $\log p_0$ directly for numerical stability.

The same term also clarifies the von Mises-Fisher interpretation of our scoring rule \cite{banerjee2005clustering}. A vMF density on $S^{d-1}$ has the form

\begin{equation}
f(u\mid \mu,\kappa)=c_d(\kappa)\exp(\kappa \mu^\top u),
\label{eq:app_vmf_density}
\end{equation}

where $\mu$ is a unit mean direction and $\kappa\ge 0$ is the concentration parameter. When $\kappa=0$, the vMF distribution reduces to the uniform distribution on the sphere. If each active reference direction is written generically as $\mathbf{r}_k$, with $\mathbf{r}_k=\boldsymbol{\psi}_k$ for base classes and $\mathbf{r}_k=\mu_k(t)$ for novel prototypes, then the log posterior of assigning $u_t$ to prototype $k$ is, up to an additive constant independent of $k$,

\begin{equation}
\log \pi_k(t)+\log c_d(\kappa)+\kappa \mathbf{r}_k^\top u_t.
\label{eq:app_vmf_logposterior}
\end{equation}

Since both $\kappa$ and $c_d(\kappa)$ are shared across prototypes, maximizing this posterior over $k$ is equivalent to maximizing

\begin{equation}
\frac{\mathbf{r}_k^\top u_t}{T}+\log \pi_k(t),
\label{eq:app_vmf_score_equiv}
\end{equation}

which is the same scoring form as $s_k(u_t)$ in the main paper once the active reference direction is instantiated as $\boldsymbol{\psi}_k$ or $\mu_k(t)$. Likewise, the log-likelihood ratio between the best prototype component and the uniform background is

\begin{equation}
\max_k\left[\frac{\mathbf{r}_k^\top u_t}{T}+\log c_d(1/T)-\log p_0\right].
\label{eq:app_vmf_llr}
\end{equation}

The term $\log c_d(1/T)$ is constant once $T$ is fixed, so it can be absorbed into the threshold. This yields the simplified birth statistic used in the paper,

\begin{equation}
\Lambda(u_t)=\max_k\frac{\mathbf{r}_k^\top u_t}{T}-\log p_0,
\label{eq:app_vmf_birth_stat}
\end{equation}

which preserves the same decision ordering while keeping the expression concise. In this sense, $p_0$ is not an ad hoc offset. It is the natural background model obtained when directional concentration vanishes.

\subsection{Novel Prototype Statistics for Attach-versus-Create}
\label{app:attach}
This subsection expands the attach-versus-create stage in \mainsection{subsec:inference}, especially the quantities behind \mainequation{eq:attach_score} and \mainequation{eq:attach_decision}. When a sample enters the novel-discovery regime, we compare it against the currently active novel prototypes rather than creating a new cluster immediately. If no novel prototype exists yet, i.e. $K_t=K_{\text{base}}$, the current sample directly creates prototype $K_t+1$; the definitions below apply when the active novel set is non-empty. Let the active novel set at time $t$ be

\begin{equation}
\mathcal{N}_t=\{K_{\text{base}}+1,\dots,K_t\}.
\label{eq:app_attach_novel_set}
\end{equation}

For each novel prototype $k\in\mathcal{N}_t$, let $n_k(t)$ be its cluster size and define the running resultant vector

\begin{equation}
R_k(t)=\sum_{\tau<t:\hat y_\tau=k}u_\tau.
\label{eq:app_attach_resultant}
\end{equation}

The corresponding unit prototype direction is

\begin{equation}
\mu_k(t)=\frac{R_k(t)}{\|R_k(t)\|_2}.
\label{eq:app_attach_mu}
\end{equation}

To characterize how concentrated the assigned samples are around that direction, we first define the mean resultant length

\begin{equation}
\bar r_k(t)=\frac{\|R_k(t)\|_2}{n_k(t)},
\label{eq:app_attach_mean_resultant}
\end{equation}

and then use the standard moment-based approximation of the vMF concentration parameter \cite{sra2012short}

\begin{equation}
\kappa_k(t)=
\frac{\bar r_k(t)\big(d-\bar r_k^2(t)\big)}{1-\bar r_k^2(t)}
\cdot
\frac{n_k(t)-1}{n_k(t)+1}.
\label{eq:app_attach_kappa}
\end{equation}

The shrinkage factor $(n_k(t)-1)/(n_k(t)+1)$ makes very small novel clusters more conservative. Using these quantities, the attach score in \mainequation{eq:attach_score} is

\begin{equation}
a_k(u_t)=\log n_k(t)+\kappa_k(t)\mu_k(t)^\top u_t-\log p_0.
\label{eq:app_attach_score}
\end{equation}

Let

\begin{equation}
k_t^\dagger=\arg\max_{k\in\mathcal{N}_t}a_k(u_t).
\label{eq:app_attach_argmax}
\end{equation}

If $a_{k_t^\dagger}(u_t)\ge \tau_{\text{create}}$, the sample is attached to prototype $k_t^\dagger$; otherwise a new prototype $K_t+1$ is created. Once the decision is made, an existing novel prototype is updated by

\begin{equation}
n_{\hat y_t}(t+1)=n_{\hat y_t}(t)+1,
\qquad
R_{\hat y_t}(t+1)=R_{\hat y_t}(t)+u_t,
\label{eq:app_attach_existing_update}
\end{equation}

followed by

\begin{equation}
\mu_{\hat y_t}(t+1)=\frac{R_{\hat y_t}(t+1)}{\|R_{\hat y_t}(t+1)\|_2}.
\label{eq:app_attach_existing_mu_update}
\end{equation}

If a new prototype is born, we initialize

\begin{equation}
n_{K_t+1}(t+1)=1,\qquad
R_{K_t+1}(t+1)=u_t,\qquad
\mu_{K_t+1}(t+1)=u_t.
\label{eq:app_attach_new_init}
\end{equation}

\subsection{Support-Spread Shrinkage and Dynamic Birth Tightening}
\label{app:birth}
This subsection expands the support-spread shrinkage and online birth-threshold refinement summarized in \mainsection{subsec:adaptive_boundary}. We denote by $\tau_{\mathrm{birth}}^{\mathrm{sup}}$ the effective support-based birth prior used to initialize online inference. In implementation, it is obtained by applying a support-spread shrinkage to the raw support-only estimate $\tau_{\text{birth}}$:

\begin{equation}
\tau_{\mathrm{birth}}^{\mathrm{sup}}
=
\tau_{\mathrm{birth}}
-\frac{c_{\mathrm{spread}}\sigma_{\mathrm{pos}}}{T},
\label{eq:app_birth_sup_shrink}
\end{equation}

where $\sigma_{\mathrm{pos}}$ is the spread of positive support compatibility scores. This makes the birth boundary more conservative and reduces base-class fragmentation.

As the stream evolves, we further refine the birth threshold dynamically using only mature novel prototypes. Let $n_{\text{base}}^{\text{med}}$ be the median support size of base classes.

Let the set of mature novel prototypes at time $t$ be

\begin{equation}
\mathcal{M}_t
=
\left\{k\in\mathcal{N}_t:\ n_k(t)\ge \mathrm{round}\!\big((n_{\text{base}}^{\text{med}})^\beta\big)\right\}.
\label{eq:app_birth_mature_set}
\end{equation}

where $\beta$ is the micro-prototype exponent. For each mature novel prototype, we define its self-explanation strength as

\begin{equation}
\lambda_{k}^{\text{self}}(t)=
\frac{n_k(t)-1}{n_k(t)+1}
\cdot
\frac{\|R_k(t)\|_2}{n_k(t)}
\cdot
\frac{1}{T}
-\log p_0.
\label{eq:app_birth_self}
\end{equation}

When $|\mathcal{M}_t|<2$, we skip the bank statistic, set $\eta_t=0$, and keep $\tau_{\text{birth}}^{(t)}=\tau_{\mathrm{birth}}^{\mathrm{sup}}$. Otherwise, from the set $\{\lambda_{k}^{\text{self}}(t)\}_{k\in\mathcal{M}_t}$ we build a robust lower fence

\begin{equation}
\tau_{\text{birth}}^{\text{bank}}(t)
=
\mathrm{median}\!\left(\{\lambda_{k}^{\text{self}}(t)\}\right)
-
\mathrm{MAD}\!\left(\{\lambda_{k}^{\text{self}}(t)\}\right).
\label{eq:app_birth_bank}
\end{equation}

Let
\[
m_t=\mathrm{median}\!\left(\{n_k(t):k\in\mathcal{M}_t\}\right)
\]
\noindent denote the median size of mature novel prototypes. We then compute a mixing coefficient

\begin{equation}
\eta_t=\frac{m_t}{m_t+n_{\text{base}}^{\text{med}}},
\label{eq:app_birth_eta}
\end{equation}

The effective birth threshold is

\begin{equation}
\tau_{\text{birth}}^{(t)}=
\min\!\left\{
\tau_{\mathrm{birth}}^{\mathrm{sup}},
\ (1-\eta_t)\tau_{\mathrm{birth}}^{\mathrm{sup}}+\eta_t\tau_{\text{birth}}^{\text{bank}}(t)
\right\}.
\label{eq:app_birth_dynamic}
\end{equation}

This rule keeps the support-based static boundary as the upper bound. When fewer than two mature novel prototypes are available, the update reduces to the support-based prior. Once stable novel prototypes have already emerged, the model is allowed to tighten the birth boundary moderately according to the state of the current memory bank. This reduces excessive late-stage prototype creation while keeping the decision process anchored to the support set.

\section{Scope and Limitations}
\label{sec:app_limitations}

\textbf{Support-set calibration depends on the quality of the support set.} In \method{}, the labeled support set provides the only structured evidence available before the stream begins. The calibration stage is built entirely on this support data, so its reliability depends on how well the support set reflects the geometry of the base classes. When the support set is extremely small, highly imbalanced, or only weakly separated, the proxy responses become less informative, and the initialization of routing, birth, and attach-versus-create decisions can become less stable. In that sense, the calibration is more reliable when the support set offers a reasonably clear view of the base-class structure.

\textbf{Single-pass online discovery remains path-dependent.} Once a sample has been routed, attached, or used to create a new prototype, the dynamic prototype memory is updated immediately and earlier assignments are not revisited. An inaccurate early birth or attachment can therefore influence later decisions through the evolving memory state and the subsequent adjustment of the birth boundary. The routing gate, support-spread shrinkage, and mature-prototype tightening are introduced to reduce this sensitivity, yet some dependence on early ambiguous observations remains, especially when stable novel structure emerges only gradually.

\ifdefined\PACOCombinedAppendix
\else
\bibliographystyle{ACM-Reference-Format}
\bibliography{sample-base}

@String{Computer = "{IEEE} Computer" }

@String{Springer = "Springer-Verlag" }

@inproceedings{11,
  title={Densely connected convolutional networks},
  author={Huang, Gao and Liu, Zhuang and Van Der Maaten, Laurens and Weinberger, Kilian Q},
  booktitle={Proceedings of the IEEE conference on computer vision and pattern recognition},
  pages={4700--4708},
  year={2017}
}

@article{22,
  title={An image is worth 16x16 words: Transformers for image recognition at scale},
  author={Dosovitskiy, Alexey},
  journal={arXiv preprint arXiv:2010.11929},
  year={2020}
}

@inproceedings{33,
  title={Deep residual learning for image recognition},
  author={He, Kaiming and Zhang, Xiangyu and Ren, Shaoqing and Sun, Jian},
  booktitle={Proceedings of the IEEE conference on computer vision and pattern recognition},
  pages={770--778},
  year={2016}
}

@inproceedings{NCD,
  title={Learning to discover novel visual categories via deep transfer clustering},
  author={Han, Kai and Vedaldi, Andrea and Zisserman, Andrew},
  booktitle={Proceedings of the IEEE/CVF international conference on computer vision},
  pages={8401--8409},
  year={2019}
}

@inproceedings{2022GCD,
  title={Generalized category discovery},
  author={Vaze, Sagar and Han, Kai and Vedaldi, Andrea and Zisserman, Andrew},
  booktitle={Proceedings of the IEEE/CVF conference on computer vision and pattern recognition},
  pages={7492--7501},
  year={2022}
}

@article{rastegar2023learn,
  title={Learn to categorize or categorize to learn? self-coding for generalized category discovery},
  author={Rastegar, Sarah and Doughty, Hazel and Snoek, Cees},
  journal={Advances in Neural Information Processing Systems},
  volume={36},
  pages={72794--72818},
  year={2023}
}

@inproceedings{zhao2023learning,
  title={Learning semi-supervised gaussian mixture models for generalized category discovery},
  author={Zhao, Bingchen and Wen, Xin and Han, Kai},
  booktitle={Proceedings of the IEEE/CVF international conference on computer vision},
  pages={16623--16633},
  year={2023}
}

@inproceedings{SMILE,
  title={On-the-fly category discovery},
  author={Du, Ruoyi and Chang, Dongliang and Liang, Kongming and Hospedales, Timothy and Song, Yi-Zhe and Ma, Zhanyu},
  booktitle={Proceedings of the IEEE/CVF Conference on Computer Vision and Pattern Recognition},
  pages={11691--11700},
  year={2023}
}

@article{PHE,
  title={Prototypical hash encoding for on-the-fly fine-grained category discovery},
  author={Zheng, Haiyang and Pu, Nan and Li, Wenjing and Sebe, Nicu and Zhong, Zhun},
  journal={Advances in Neural Information Processing Systems},
  volume={37},
  pages={101428--101455},
  year={2024}
}

@inproceedings{DiffGRE,
  title={Generate, refine, and encode: Leveraging synthesized novel samples for on-the-fly fine-grained category discovery},
  author={Liu, Xiao and Pu, Nan and Zheng, Haiyang and Li, Wenjing and Sebe, Nicu and Zhong, Zhun},
  booktitle={Proceedings of the IEEE/CVF International Conference on Computer Vision},
  pages={1078--1087},
  year={2025}
}

@inproceedings{tau1,
  title={Learning placeholders for open-set recognition},
  author={Zhou, Da-Wei and Ye, Han-Jia and Zhan, De-Chuan},
  booktitle={Proceedings of the IEEE/CVF conference on computer vision and pattern recognition},
  pages={4401--4410},
  year={2021}
}

@inproceedings{tau2,
  title={Learning for Transductive Threshold Calibration in Open-World Recognition},
  author={Zhang, Qin and An, Dongsheng and Xiao, Tianjun and He, Tong and Tang, Qingming and Wu, Ying Nian and Tighe, Joseph and Xing, Yifan},
  booktitle={Proceedings of the IEEE/CVF Conference on Computer Vision and Pattern Recognition},
  pages={17097--17106},
  year={2024}
}

@article{tau3,
  title={Clustering data streams based on shared density between micro-clusters},
  author={Hahsler, Michael and Bola{\~n}os, Matthew},
  journal={IEEE transactions on knowledge and data engineering},
  volume={28},
  number={6},
  pages={1449--1461},
  year={2016},
  publisher={IEEE}
}

@inproceedings{tau4,
  title={Density-based clustering over an evolving data stream with noise},
  author={Cao, Feng and Estert, Martin and Qian, Weining and Zhou, Aoying},
  booktitle={Proceedings of the 2006 SIAM international conference on data mining},
  pages={328--339},
  year={2006},
  organization={SIAM}
}

@inproceedings{DINO,
  title={Emerging properties in self-supervised vision transformers},
  author={Caron, Mathilde and Touvron, Hugo and Misra, Ishan and J{\'e}gou, Herv{\'e} and Mairal, Julien and Bojanowski, Piotr and Joulin, Armand},
  booktitle={Proceedings of the IEEE/CVF international conference on computer vision},
  pages={9650--9660},
  year={2021}
}

@inproceedings{arc,
  title={Arcface: Additive angular margin loss for deep face recognition},
  author={Deng, Jiankang and Guo, Jia and Xue, Niannan and Zafeiriou, Stefanos},
  booktitle={Proceedings of the IEEE/CVF conference on computer vision and pattern recognition},
  pages={4690--4699},
  year={2019}
}

@article{li2019tutorial,
  title={A tutorial on Dirichlet process mixture modeling},
  author={Li, Yuelin and Schofield, Elizabeth and G{\"o}nen, Mithat},
  journal={Journal of mathematical psychology},
  volume={91},
  pages={128--144},
  year={2019},
  publisher={Elsevier}
}

@article{banerjee2005clustering,
  title={Clustering on the Unit Hypersphere using von Mises-Fisher Distributions.},
  author={Banerjee, Arindam and Dhillon, Inderjit S and Ghosh, Joydeep and Sra, Suvrit and Ridgeway, Greg},
  journal={Journal of Machine Learning Research},
  volume={6},
  number={9},
  year={2005}
}

@article{sra2012short,
  title={A short note on parameter approximation for von Mises-Fisher distributions: and a fast implementation of I s (x)},
  author={Sra, Suvrit},
  journal={Computational Statistics},
  volume={27},
  number={1},
  pages={177--190},
  year={2012},
  publisher={Springer}
}

@book{hartigan1975clustering,
  title={Clustering algorithms},
  author={Hartigan, John A},
  year={1975},
  publisher={John Wiley \& Sons, Inc.}
}

@article{RankStat,
  title={Autonovel: Automatically discovering and learning novel visual categories},
  author={Han, Kai and Rebuffi, Sylvestre-Alvise and Ehrhardt, Sebastien and Vedaldi, Andrea and Zisserman, Andrew},
  journal={IEEE Transactions on Pattern Analysis and Machine Intelligence},
  volume={44},
  number={10},
  pages={6767--6781},
  year={2021},
  publisher={IEEE}
}

@inproceedings{WTAjia2021joint,
  title={Joint representation learning and novel category discovery on single-and multi-modal data},
  author={Jia, Xuhui and Han, Kai and Zhu, Yukun and Green, Bradley},
  booktitle={Proceedings of the IEEE/CVF international conference on computer vision},
  pages={610--619},
  year={2021}
}

@inproceedings{MLDGli2018learning,
  title={Learning to generalize: Meta-learning for domain generalization},
  author={Li, Da and Yang, Yongxin and Song, Yi-Zhe and Hospedales, Timothy},
  booktitle={Proceedings of the AAAI conference on artificial intelligence},
  volume={32},
  number={1},
  year={2018}
}

@article{CUB,
  title={The caltech-ucsd birds-200-2011 dataset},
  author={Wah, Catherine and Branson, Steve and Welinder, Peter and Perona, Pietro and Belongie, Serge},
  year={2011},
  publisher={california institute of technology}
}

@inproceedings{scars,
  title={3d object representations for fine-grained categorization},
  author={Krause, Jonathan and Stark, Michael and Deng, Jia and Fei-Fei, Li},
  booktitle={Proceedings of the IEEE international conference on computer vision workshops},
  pages={554--561},
  year={2013}
}

@inproceedings{pets,
  title={Cats and dogs},
  author={Parkhi, Omkar M and Vedaldi, Andrea and Zisserman, Andrew and Jawahar, CV},
  booktitle={2012 IEEE conference on computer vision and pattern recognition},
  pages={3498--3505},
  year={2012},
  organization={IEEE}
}

@inproceedings{food,
  title={Food-101--mining discriminative components with random forests},
  author={Bossard, Lukas and Guillaumin, Matthieu and Van Gool, Luc},
  booktitle={European conference on computer vision},
  pages={446--461},
  year={2014},
  organization={Springer}
}

@article{cifar,
  title={Learning multiple layers of features from tiny images},
  author={Krizhevsky, Alex and Hinton, Geoffrey and others},
  year={2009},
  publisher={Toronto, ON, Canada}
}

@article{2015imagenet,
  title={Imagenet large scale visual recognition challenge},
  author={Russakovsky, Olga and Deng, Jia and Su, Hao and Krause, Jonathan and Satheesh, Sanjeev and Ma, Sean and Huang, Zhiheng and Karpathy, Andrej and Khosla, Aditya and Bernstein, Michael and others},
  journal={International journal of computer vision},
  volume={115},
  number={3},
  pages={211--252},
  year={2015},
  publisher={Springer}
}

@article{sync,
  title={Language-assisted Feature Representation and Lightweight Active Learning For On-the-Fly Category Discovery},
  author={Banerjee, Anwesha and Biswas, Soma},
  year={2025},
  journal={Transactions on Machine Learning Research}
}

@inproceedings{AGE,
  title={Adaptive Gaussian Expansion for On-the-fly Category Discovery},
  author={Li, Chunming and Wang, Shidong and Zhang, Haofeng},
  year={2025},
  booktitle={The Fourteenth International Conference on Learning Representations}
}

@inproceedings{wu2023metagcd,
  title={Metagcd: Learning to continually learn in generalized category discovery},
  author={Wu, Yanan and Chi, Zhixiang and Wang, Yang and Feng, Songhe},
  booktitle={Proceedings of the IEEE/CVF international conference on computer vision},
  pages={1655--1665},
  year={2023}
}

@inproceedings{wen2023parametric,
  title={Parametric classification for generalized category discovery: A baseline study},
  author={Wen, Xin and Zhao, Bingchen and Qi, Xiaojuan},
  booktitle={Proceedings of the IEEE/CVF international conference on computer vision},
  pages={16590--16600},
  year={2023}
}

@article{vaze2023no,
  title={No representation rules them all in category discovery},
  author={Vaze, Sagar and Vedaldi, Andrea and Zisserman, Andrew},
  journal={Advances in Neural Information Processing Systems},
  volume={36},
  pages={19962--19989},
  year={2023}
}

@inproceedings{liu2025collaborative,
  title={Collaborative Cloud-edge Generalized Category Discovery},
  author={Liu, Yingbing and Ma, Fei and Wu, Yanan and Zuo, Xinxin and Zhang, Fan and Wang, Yang},
  booktitle={Proceedings of the 33rd ACM International Conference on Multimedia},
  pages={535--543},
  year={2025}
}

@inproceedings{miller2021class,
  title={Class anchor clustering: A loss for distance-based open set recognition},
  author={Miller, Dimity and Sunderhauf, Niko and Milford, Michael and Dayoub, Feras},
  booktitle={Proceedings of the IEEE/CVF Winter Conference on Applications of Computer Vision},
  pages={3570--3578},
  year={2021}
}

@inproceedings{NCD1,
  title={Boosting novel category discovery over domains with soft contrastive learning and all in one classifier},
  author={Zang, Zelin and Shang, Lei and Yang, Senqiao and Wang, Fei and Sun, Baigui and Xie, Xuansong and Li, Stan Z},
  booktitle={Proceedings of the IEEE/CVF International Conference on Computer Vision},
  pages={11858--11867},
  year={2023}
}

@article{NCD2,
  title={Novel visual category discovery with dual ranking statistics and mutual knowledge distillation},
  author={Zhao, Bingchen and Han, Kai},
  journal={Advances in Neural Information Processing Systems},
  volume={34},
  pages={22982--22994},
  year={2021}
}

@article{NCD3,
  title={Novel class discovery for long-tailed recognition},
  author={Zhang, Chuyu and Xu, Ruijie and He, Xuming},
  journal={arXiv preprint arXiv:2308.02989},
  year={2023}
}

@inproceedings{tau5,
  title={Towards open set deep networks},
  author={Bendale, Abhijit and Boult, Terrance E},
  booktitle={Proceedings of the IEEE conference on computer vision and pattern recognition},
  pages={1563--1572},
  year={2016}
}

@inproceedings{tau7,
  title={Task-adaptive negative envision for few-shot open-set recognition},
  author={Huang, Shiyuan and Ma, Jiawei and Han, Guangxing and Chang, Shih-Fu},
  booktitle={Proceedings of the IEEE/CVF Conference on Computer Vision and Pattern Recognition},
  pages={7171--7180},
  year={2022}
}

@inproceedings{tau8,
  title={Operational open-set recognition and postmax refinement},
  author={Cruz, Steve and Rabinowitz, Ryan and G{\"u}nther, Manuel and Boult, Terrance E},
  booktitle={European Conference on Computer Vision},
  pages={475--492},
  year={2024},
  organization={Springer}
}

@inproceedings{ma1,
  title={Active generalized category discovery},
  author={Ma, Shijie and Zhu, Fei and Zhong, Zhun and Zhang, Xu-Yao and Liu, Cheng-Lin},
  booktitle={Proceedings of the IEEE/CVF conference on computer vision and pattern recognition},
  pages={16890--16900},
  year={2024}
}

@article{ma2,
  title={Happy: A debiased learning framework for continual generalized category discovery},
  author={Ma, Shijie and Zhu, Fei and Zhong, Zhun and Liu, Wenzhuo and Zhang, Xu-Yao and Liu, Cheng-Lin},
  journal={Advances in Neural Information Processing Systems},
  volume={37},
  pages={50850--50875},
  year={2024}
}

@article{ma3,
  title={Protogcd: Unified and unbiased prototype learning for generalized category discovery},
  author={Ma, Shijie and Zhu, Fei and Zhang, Xu-Yao and Liu, Cheng-Lin},
  journal={IEEE Transactions on Pattern Analysis and Machine Intelligence},
  year={2025},
  publisher={IEEE}
}

@article{vaswani2017attention,
  title={Attention is all you need},
  author={Vaswani, Ashish and Shazeer, Noam and Parmar, Niki and Uszkoreit, Jakob and Jones, Llion and Gomez, Aidan N and Kaiser, {\L}ukasz and Polosukhin, Illia},
  journal={Advances in neural information processing systems},
  volume={30},
  year={2017}
}

@article{lecun2015deep,
  title={Deep learning},
  author={LeCun, Yann and Bengio, Yoshua and Hinton, Geoffrey},
  journal={nature},
  volume={521},
  number={7553},
  pages={436--444},
  year={2015},
  publisher={Nature Publishing Group UK London}
}

@incollection{herrera2016multilabel,
  title={Multilabel classification},
  author={Herrera, Francisco and Charte, Francisco and Rivera, Antonio J and Del Jesus, Mar{\'\i}a J},
  booktitle={Multilabel Classification: Problem Analysis, Metrics and Techniques},
  pages={17--31},
  year={2016},
  publisher={Springer}
}

@inproceedings{choi2024contrastive,
  title={Contrastive mean-shift learning for generalized category discovery},
  author={Choi, Sua and Kang, Dahyun and Cho, Minsu},
  booktitle={Proceedings of the IEEE/CVF conference on computer vision and pattern recognition},
  pages={23094--23104},
  year={2024}
}

@article{wang2024sptnet,
  title={Sptnet: An efficient alternative framework for generalized category discovery with spatial prompt tuning},
  author={Wang, Hongjun and Vaze, Sagar and Han, Kai},
  journal={arXiv preprint arXiv:2403.13684},
  year={2024}
}

@inproceedings{rastegar2024selex,
  title={Selex: Self-expertise in fine-grained generalized category discovery},
  author={Rastegar, Sarah and Salehi, Mohammadreza and Asano, Yuki M and Doughty, Hazel and Snoek, Cees GM},
  booktitle={European Conference on Computer Vision},
  pages={440--458},
  year={2024},
  organization={Springer}
}

@inproceedings{pu2024federated,
  title={Federated generalized category discovery},
  author={Pu, Nan and Li, Wenjing and Ji, Xingyuan and Qin, Yalan and Sebe, Nicu and Zhong, Zhun},
  booktitle={Proceedings of the IEEE/CVF conference on computer vision and pattern recognition},
  pages={28741--28750},
  year={2024}
}

@inproceedings{liu2024novel,
  title={Novel class discovery for ultra-fine-grained visual categorization},
  author={Liu, Yu and Cai, Yaqi and Jia, Qi and Qiu, Binglin and Wang, Weimin and Pu, Nan},
  booktitle={Proceedings of the IEEE/CVF Conference on Computer Vision and Pattern Recognition},
  pages={17679--17688},
  year={2024}
}

@inproceedings{zhao2024labeled,
  title={Labeled data selection for category discovery},
  author={Zhao, Bingchen and Lang, Nico and Belongie, Serge and Aodha, Oisin Mac},
  booktitle={European Conference on Computer Vision},
  pages={201--218},
  year={2024},
  organization={Springer}
}

@inproceedings{liu2025hyperbolic,
  title={Hyperbolic category discovery},
  author={Liu, Yuanpei and He, Zhenqi and Han, Kai},
  booktitle={Proceedings of the Computer Vision and Pattern Recognition Conference},
  pages={9891--9900},
  year={2025}
}

@inproceedings{cao2025allgcd,
  title={AllGCD: Leveraging All Unlabeled Data for Generalized Category Discovery},
  author={Cao, Xinzi and Chen, Ke and Yang, Feidiao and Zheng, Xiawu and Tian, Yonghong and Lu, Yutong},
  booktitle={Proceedings of the IEEE/CVF International Conference on Computer Vision},
  pages={3293--3303},
  year={2025}
}

@inproceedings{peng2025mos,
  title={Mos: Modeling object-scene associations in generalized category discovery},
  author={Peng, Zhengyuan and Ma, Jinpeng and Sun, Zhimin and Yi, Ran and Song, Haichuan and Tan, Xin and Ma, Lizhuang},
  booktitle={Proceedings of the Computer Vision and Pattern Recognition Conference},
  pages={15118--15128},
  year={2025}
}

@inproceedings{wang2025get,
  title={Get: Unlocking the multi-modal potential of clip for generalized category discovery},
  author={Wang, Enguang and Peng, Zhimao and Xie, Zhengyuan and Yang, Fei and Liu, Xialei and Cheng, Ming-Ming},
  booktitle={Proceedings of the Computer Vision and Pattern Recognition Conference},
  pages={20296--20306},
  year={2025}
}

@inproceedings{zhang2025less,
  title={Less attention is more: Prompt transformer for generalized category discovery},
  author={Zhang, Wei and Zhang, Baopeng and Teng, Zhu and Luo, Wenxin and Zou, Junnan and Fan, Jianping},
  booktitle={Proceedings of the Computer Vision and Pattern Recognition Conference},
  pages={30322--30331},
  year={2025}
}

@inproceedings{rathore2025domain,
  title={When Domain Generalization meets Generalized Category Discovery: An Adaptive Task-Arithmetic Driven Approach},
  author={Rathore, Vaibhav and Dutta, Saikat and Mehrotra, Sarthak and Kira, Zsolt and Banerjee, Biplab and others},
  booktitle={Proceedings of the Computer Vision and Pattern Recognition Conference},
  pages={4905--4915},
  year={2025}
}

@article{cao2021open,
  title={Open-world semi-supervised learning},
  author={Cao, Kaidi and Brbic, Maria and Leskovec, Jure},
  journal={arXiv preprint arXiv:2102.03526},
  year={2021}
}

@inproceedings{rizve2022openldn,
  title={Openldn: Learning to discover novel classes for open-world semi-supervised learning},
  author={Rizve, Mamshad Nayeem and Kardan, Navid and Khan, Salman and Shahbaz Khan, Fahad and Shah, Mubarak},
  booktitle={European Conference on Computer Vision},
  pages={382--401},
  year={2022},
  organization={Springer}
}

@article{ouldnoughi2023clip,
  title={Clip-gcd: Simple language guided generalized category discovery},
  author={Ouldnoughi, Rabah and Kuo, Chia-Wen and Kira, Zsolt},
  journal={arXiv preprint arXiv:2305.10420},
  year={2023}
}

@inproceedings{zheng2024textual,
  title={Textual knowledge matters: Cross-modality co-teaching for generalized visual class discovery},
  author={Zheng, Haiyang and Pu, Nan and Li, Wenjing and Sebe, Nicu and Zhong, Zhun},
  booktitle={European Conference on Computer Vision},
  pages={41--58},
  year={2024},
  organization={Springer}
}

@inproceedings{shi2024unified,
  title={A unified knowledge transfer network for generalized category discovery},
  author={Shi, Wenkai and An, Wenbin and Tian, Feng and Chen, Yan and Wu, Yaqiang and Wang, Qianying and Chen, Ping},
  booktitle={Proceedings of the AAAI Conference on Artificial Intelligence},
  volume={38},
  number={17},
  pages={18961--18969},
  year={2024}
}

@article{wang2024hilo,
  title={Hilo: A learning framework for generalized category discovery robust to domain shifts},
  author={Wang, Hongjun and Vaze, Sagar and Han, Kai},
  journal={arXiv preprint arXiv:2408.04591},
  year={2024}
}

\end{document}
\fi

\end{document}